\documentclass[a4paper,fleqn]{article}
\usepackage[margin=0.9in]{geometry}

\usepackage[square,numbers]{natbib}
\setcitestyle{numbers}

\usepackage{url}

\usepackage{breakurl}
\usepackage[breaklinks,pagebackref]{hyperref}

\usepackage{multirow}
\usepackage{times}
\usepackage[group-separator={,},group-minimum-digits=4]{siunitx}
\usepackage{epsfig}
\usepackage{graphicx}
\usepackage{amsmath}
\usepackage{amssymb}
\usepackage{booktabs}

\usepackage{comment}
\usepackage[mathscr]{euscript}
\usepackage{bbm}
\usepackage{soul}
\usepackage[normalem]{ulem}

\usepackage{xspace}
\usepackage{subfig}
\usepackage{svg}
\usepackage{algorithm}
\usepackage[noend]{algpseudocode}
\usepackage{xcolor}

\newcommand{\x}{\boldsymbol{x}}

\newcommand{\inn}[0]{\mathtt{INN}_{\boldsymbol \Theta}}

\newcommand{\f}[1]{f_{\theta_{#1}}^{(#1)}}
\newcommand{\qpnn}[0]{\mathtt{QPNN}_{\boldsymbol {\Tilde \Theta}}}
\newcommand{\qmpnn}[0]{\mathtt{QMPNN}_{\boldsymbol {\Tilde \Theta}}}
\newcommand{\bmpnn}[0]{\mathtt{BMPNN}_{\boldsymbol {\Theta}}}

\newcommand\y[0]{\boldsymbol{y}}
\newcommand{\bPhi}[0]{\boldsymbol \Phi}

\newcommand{\Qmpnn}[0]{\textsc{Qmpnn}\xspace}
\newcommand{\Bmpnn}[0]{\textsc{Bmpnn}\xspace}
\newcommand{\Qpnn}[0]{\textsc{Qpnn}\xspace}
\newcommand{\Bpnn}[0]{\textsc{Bpnn}\xspace}

\newcommand{\comp}[0]{\textsc{Comp}\xspace}
\newcommand{\dorf}[0]{\textsc{Dorfman}\xspace}
\newcommand{\individual}[0]{\textsc{Individual}\xspace}
\newcommand{\ncomp}[0]{\textsc{NComp}\xspace}

\newcommand{\milp}[0]{\textsc{Mip}\xspace}
\newcommand{\classo}[0]{\textsc{CLasso}\xspace}

\newcommand{\nlpd}[0]{\mathtt{NLPD}}

\newcommand{\dorfod}[1]{\textsc{Dorfman-{#1}-OD}\xspace}

\newcommand{\classood}[0]{\textsc{CLasso-OD}\xspace}
\newcommand{\compod}[0]{\textsc{Comp-OD}\xspace}
\newcommand{\ncompod}[1]{\textsc{NComp-{#1}-OD}\xspace}
\newcommand{\individualod}[0]{\textsc{Individual-OD}\xspace}
\newcommand{\downsampleod}[0]{\textsc{Downsample-OD}\xspace}

\makeatletter
\def\blfootnote{\xdef\@thefnmark{}\@footnotetext}
\makeatother

\begin{document}
\let\WriteBookmarks\relax
\def\floatpagepagefraction{1}
\def\textpagefraction{.001}

\title{
Efficient Neural Network based Classification and Outlier Detection for Image Moderation using Compressed Sensing and Group Testing
}

\author{
Sabyasachi Ghosh\\
\texttt{sghosh@cse.iitb.ac.in}
\and
Sanyam Saxena\\
\texttt{sanyamsaxena@cse.iitb.ac.in}
\and
Ajit Rajwade\thanks{Corresponding Author}\\
\texttt{ajitvr@cse.iitb.ac.in}
}

\date{Department of Computer Science and Engineering,\\ IIT Bombay, Mumbai 400076, India}

\maketitle

\begin{abstract}
Popular social media platforms which allow users to upload image content employ neural network based image moderation engines to classify images as having potentially objectionable or dangerous content (such as images depicting weapons, drugs, nudity, etc).
As millions of images are shared everyday, such image moderation engines must answer a large number of queries with heavy computational cost, even though the actual number of images with objectionable content is usually a tiny fraction of the total number.
Inspired by recent work on Neural Group Testing, we propose an approach which exploits this fact to reduce the overall computational cost of such neural network based image moderation engines using the technique of Compressed Sensing (CS).
We present the \emph{quantitative matrix-pooled neural network} (\Qmpnn), which takes as input $n$ images, and a $m \times n$ binary pooling matrix with $m < n$, whose rows indicate $m$ pools of images i.e. selections of $r$ images out of $n$.
The \Qmpnn efficiently outputs the product of this matrix with the unknown sparse binary vector representing the classification of each image as objectionable or non-objectionable, i.e., it outputs the number of objectionable images in each pool.
If the matrix obeys certain properties as required by CS theory, this compressed representation can then be decoded using CS algorithms to predict which input images were objectionable.
The computational cost of running the \Qmpnn and the CS decoding algorithms is significantly lower than the cost of using a neural network with the same number of parameters separately on each image to classify the images, which we demonstrate via extensive experiments.
Our technique is inherently resilient to moderate levels of errors in the prediction from the \Qmpnn.
Furthermore, we present pooled deep outlier detection, which 
brings CS and group testing techniques to deep outlier detection, to provide for the important case when the objectionable images do not belong to a set of pre-defined classes.
This technique is designed to enable efficient automated moderation of off-topic images shared on topical forums dedicated to sharing images of a certain single class, many of which are currently human-moderated.
\end{abstract}

\section{Introduction}
\label{sec:intro}
\blfootnote{A list of abbreviations used in this paper is provided in Table~\ref{tab:abbreviations}}
Compressed sensing (CS) has been a very extensively studied branch of signal/image processing, which involves acquiring signals/images directly in compressed form as opposed to performing compression post acquisition. Consider a possibly noisy vector $\boldsymbol{y} \in \mathbb{R}^m$ of measurements of the signal $\boldsymbol{x} \in \mathbb{R}^n$ with $m \ll n$, where the measurements are acquired via a sensing matrix $\boldsymbol{\Phi} \in \mathbb{R}^{m \times n}$ (implemented in hardware). Then we have the relationship:
\begin{equation}
    \boldsymbol{y} = \boldsymbol{\Phi x} + \boldsymbol{\eta},
    \label{eq:forward}
\end{equation}    
where $\boldsymbol{\eta} \in \mathbb{R}^m$ is a vector of i.i.d. noise values. CS theory \cite{Candes2008a,THW2015} states that under two conditions, $\boldsymbol{x}$ can be stably and robustly recovered from $\boldsymbol{y}, \boldsymbol{\Phi}$, with rigorous theoretical guarantees, by solving convex optimization problems such as the \textsc{Lasso} \cite{THW2015}, given as follows:
\begin{equation}
    \boldsymbol{\hat{x}} \triangleq \textrm{argmin}_{\boldsymbol{x}} \|\boldsymbol{y} - \boldsymbol{\Phi x}\|^2 + \lambda \|\boldsymbol{x}\|_1,
    \label{eq:lasso}
\end{equation}
where $\lambda$ is a carefully chosen regularization parameter.
The two conditions are: ($\mathscr{C}1$) $\boldsymbol{x}$ should be a sufficiently sparse vector, and ($\mathscr{C}2$) no sparse vector, except for a vector of all zeroes, should lie in the null-space of $\boldsymbol{\Phi}$.
$\mathscr{C}1$ and $\mathscr{C}2$ ensure that $\boldsymbol{y}$ and $\boldsymbol{x}$ are uniquely mapped to each other via $\boldsymbol{\Phi}$ even though $m \ll n$.
$\mathscr{C}2$ is typically satisfied when the entries of $\boldsymbol{\Phi}$ belong to sub-Gaussian distributions and when $m \geq O(k \log n/k)$. If these conditions are met, then successful recovery of a vector $\boldsymbol{x}$ with at the most $k$ non-zero elements is ensured \cite{Candes2008a,THW2015}. 

Group testing (GT), also called pooled testing, is an area of information theory which is closely related to CS \cite{Gilbert2008}. Given a set of $n$ samples (for example, blood/urine samples) that need to be tested for a \emph{rare} disease, GT replaces tests on the individual samples by tests on \textit{pools} (also called \textit{groups}), where each pool is created using a subset of the $n$ samples. Given the test results on each pool (as to whether or not the pool contains one or more diseased samples), the aim is to infer the status of the individual $n$ samples. This approach dates back to the classical work of Dorfman \cite{Dorfman1943,Aldridge2019} and has been recently very successful in saving resources in COVID-19 RT-PCR testing \cite{Ghosh2021,Shental_sciencemag}. In some cases, side information in the form of contact tracing matrices has also been used to further enhance the results \cite{Hasaninasab2022,Goenka2021,Goenka2021_icassp}. If the test results report the \emph{number} of defective/diseased samples in a pool instead of just a binary result, it is referred to as quantitative group testing (QGT) \cite{Scarlett2017}. 

\textbf{Image moderation} is the process of examining image content to determine whether it depicts any objectionable content or violates copyright issues. In this paper, we are concerned with semantic image moderation to determine whether the image \emph{content} is objectionable, for example depiction of violence (such as images of weapons), or whether it is off-topic for the chosen forum (such as images of a tennis game being shared on an online forum for baseball). 

Manually screening the millions of images shared on online forums like Facebook, Instagram, etc. is very tedious. There exist many commercial solutions for automated image moderation (Amazon, Azure, Picpurify, Webpurify). However, there exists only a small-sized body of academic research work in this area, mostly focused on specific categories of objectionable content, such as firearms/guns \cite{Bhatti2021}, knives \cite{Grega2016,Debnath2021} or detection of violent scenes in images/videos \cite{FreireObregon2022}. Most of these papers employ neural networks for classification, given their success in image classification tasks since \cite{Krizhevsky2012}. However large-scale neural networks require several Giga-flops of operations for a single forward pass \cite{Bianco2018} and consume considerable amounts of power as shown in \cite{Bhadar2021,Patterson2021}. Thus, methods to reduce the heavy load on moderation servers are the need of the hour. Besides reduction in power consumption, it is also helpful if the computational cost for image moderation engines is reduced significantly.

\begin{figure*}
  \centering
  \begin{minipage}[c]{.75\linewidth}
    {\includegraphics[width=\linewidth]{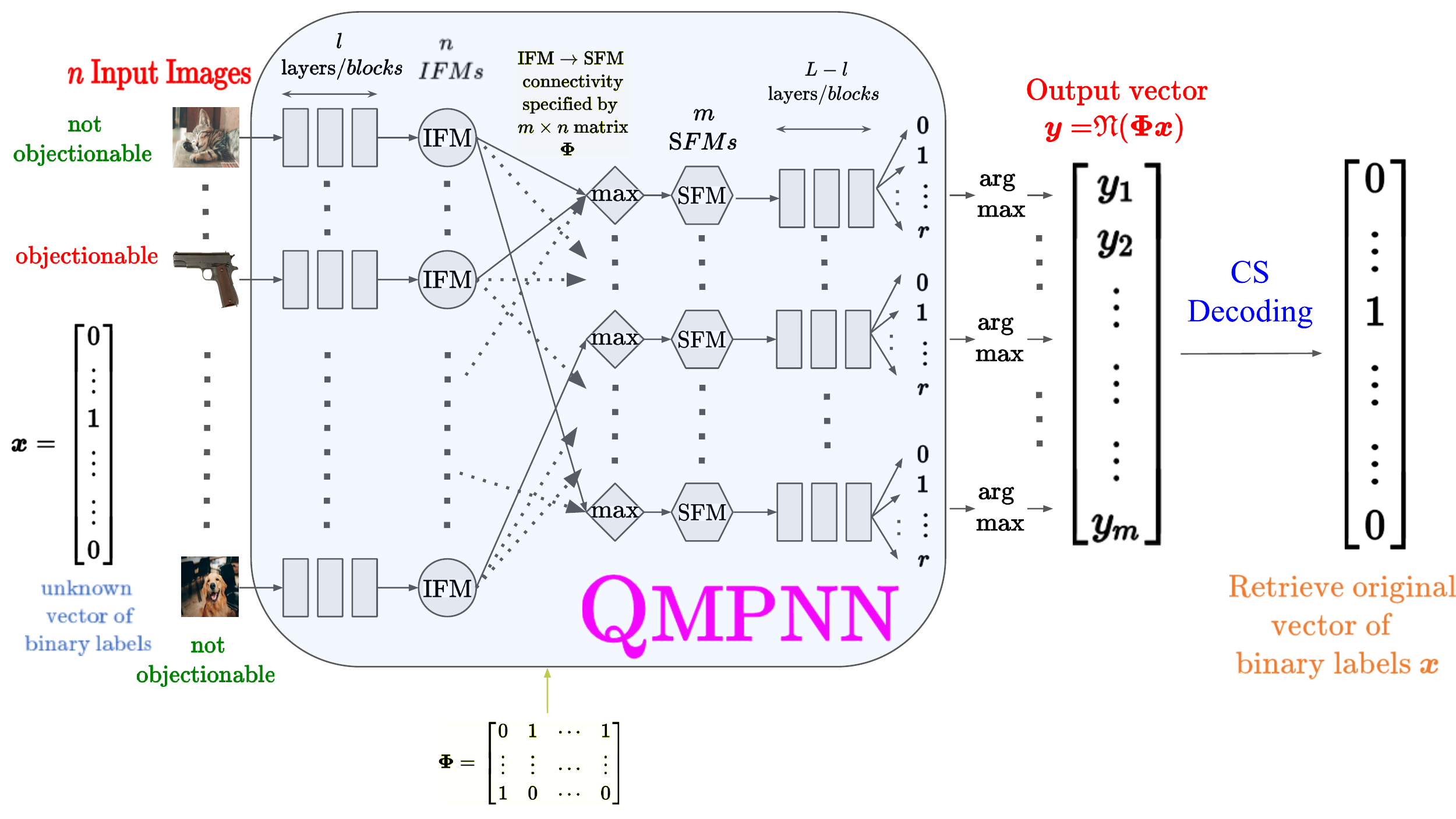}}
  \end{minipage}%
  \hfill
  \begin{minipage}[c]{.20\linewidth}
    {\includegraphics[width=.85\linewidth]{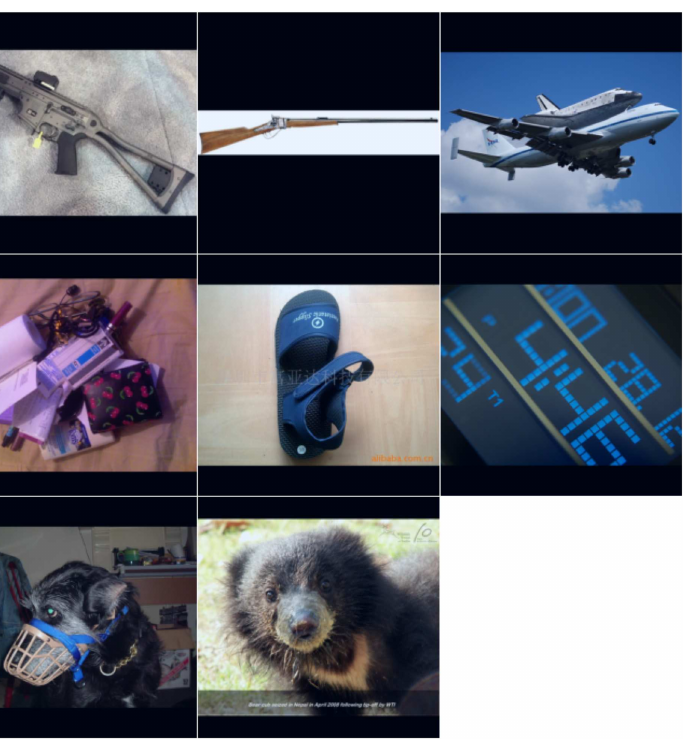}}
    {\includegraphics[width=.85\linewidth]{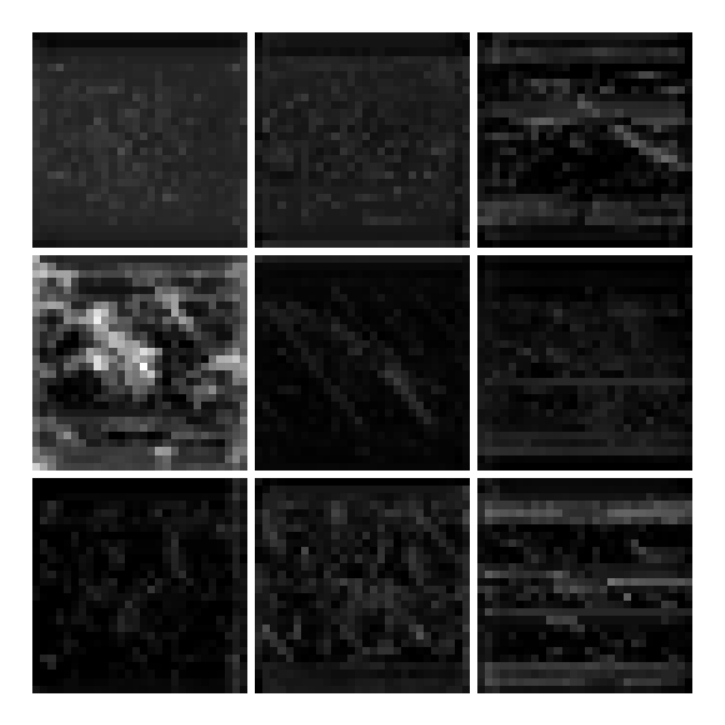}}
  \end{minipage}%
  \caption{Left: Main Components of our Image Moderation Engine; Right top: A pool of $8$ images containing $2$ images of firearms which are considered objectionable images (OIs); Right, bottom: The first $9$ channels of the Superposed Feature Map (SFM) for this pool. The SFM has size $512 \times 28\times 28$.}
    \label{fig:pooled_image_and_sfm}
\end{figure*}

\begin{table}[]
\centering
\begin{tabular}{|ll|}
\hline
\multicolumn{2}{|l|}{\textbf{List of Abbreviations}}                         \\ \hline
\textsc{Bmpnn}          & Binary Matrix-Pooled Neural Network                         \\
\textsc{Bpnn}           & Binary Pooled Neural Network                                \\
\textsc{CLasso}         & Constrained Least Absolute Shrinkage and Selection Operator \\
\textsc{Comp}           & Combinatorial Orthogonal Matching Pursuit                   \\
CS             & Compressed Sensing                                          \\
GMM            & Gaussian Mixture Model                                      \\
GT             & Group Testing                                               \\
IFM            & Intermediate Feature Map                                    \\
\textsc{Inn}            & Individual Neural Network                                   \\
\textsc{Mip}            & Mixed Integer Programming Method                            \\
\textsc{NComp}          & Noisy Combinatorial Orthogonal Matching Pursuit             \\
NLPD           & Negative Log Probability Density                            \\
OD             & Outlier Detection                                           \\
OI             & Objectionable Image                                         \\
QGT            & Quantitative Group Testing                                  \\
\textsc{Qmpnn} & Quantitative Matrix-Pooled Neural Network                   \\
\textsc{Qpnn}           & Quantitative Pooled Neural Network                          \\
SFM            & Superposed Feature Map                                      \\ \hline
\end{tabular}
\caption{List of Abbreviations}
\label{tab:abbreviations}
\end{table}

In this work, we present a CS approach to speed up image moderation.
Consider a set of $n$ images, each of which may independently have objectionable content with a small probability $p$, called the \textit{prevalence rate} of objectionable images (henceforth referred to as OIs).
Small $p$ is justified by independent reports from forums such as Facebook \cite{facebook2021transparencyreport} ($0.03-0.14\%$) or Reddit \cite{reddit2020transparencyreport} ($6\%$), where millions of images are regularly uploaded and where image or content moderation is an important requirement.
Instead of invoking a neural network separately on each image to classify it as objectionable or non-objectionable, we introduce the \textit{quantitative matrix-pooled neural network} (\textsc{Qmpnn}), which takes in a specification of $m < n$ pools of images -- each pool being a selection of $r$ out of $n$ images -- and efficiently predicts the count of OIs in these pools.
Similar to \cite{liang2021neural}, the \Qmpnn runs the first few layers of the neural network on each image of a pool and computes their Intermediate Feature Maps (IFMs), computes the superposed feature map (SFM) of the pool from its constituent IFMs, and processes only the SFM using the remaining layers of the network.
However, unlike \cite{liang2021neural}, the IFM for each image is computed only once, and all $n$ images are processed in a single forward pass to produce the $m$ pool outputs.
Such a design ensures that the computational cost of running the network on the $m$ pools is lower than that of running the network on the $n$ images individually.
The pool specification is given by an $m\times n$ binary pooling matrix. The output of the \Qmpnn is thus the product of the pooling matrix and the \emph{sparse} unknown binary vector specifying whether each image was objectionable or not.
We employ CS algorithms, which run at very little additional cost, to determine the status of each of the $n$ images from the \textsc{Qmpnn}'s results on the $m$ pools.

The above method and the one used in \cite{liang2021neural} are applicable only to classification problems, where the objectionable images belong to a single or a limited number of classes.
However, in many commonly occurring cases, the the objectionable images may belong to a large number, in fact even an unknown number of classes.
For example, on a topical forum dedicated to sharing information about tennis, any image whose content is not related to tennis can be considered objectionable.
This no longer belongs to the realm of classification, but is instead an outlier detection problem.
We extend our earlier approach of combining neural networks and CS algorithms to deal with this application via a novel technique.
Our work in this paper is inspired by that in \cite{liang2021neural} which was the first work to apply GT principles to the problem of binary classification in the presence of class imbalance. However, we significantly build up on their setup and technique, and make the following contributions (see also Sec.~\ref{subsec:differences} for more details):
\begin{enumerate}
    \item We show that the problem of classifying the original $n$ images (as OI or not) from these $m$ counts (each count belonging to a pool) can be framed as a CS problem with some additional constraints (Sec.~\ref{sec:main_approach}).
    We employ various algorithms from the CS and GT literature for this purpose (Sec.~\ref{subsec:algorithms}).
    Via an efficient implementation of our network called \textsc{Qmpnn}, we enable usage of pooling matrices in which one image can be a part of \emph{many} pools. As opposed to this, in the approach in \cite{liang2021neural}, each new pool that an image contributes to, incurs an additional cost, which makes it impractical to have an image contribute to more than two pools. Since each image contributes to many pools, an improved resilience to any errors in the network outputs can be obtained.
    \item We test our method for a wide range of prevalence rates $p$ and show that it achieves significant reduction in computation cost and end-to-end wall-clock time compared to individual moderation of images, while remaining competitive in terms of classification accuracy despite noisy outputs by the pooled neural network (Sec.~\ref{sec:exp}).
    \item We also show that our method outperforms binary group testing methods, as used in \cite{liang2021neural}, for high $p$, where quantitative information is more important (Sec.~\ref{sec:exp}).
    \item Our decoding algorithms are designed to be noise-tolerant, as opposed to existing GT algorithms used for this problem which do not handle the case of a pool falsely testing negative for the presence of an objectionable image as done in \cite{liang2021neural} (also see Sec.~\ref{sec:exp}).
    \item Finally, we present a novel \emph{pooled deep outlier detection} method for computationally efficient automatic identification of off-topic images on internet forums such as Reddit (Sec.~\ref{subsec:outlierdetection}). Moderation of such off-topic content is currently either done manually or via text-based tools \cite{jhaver2019human}. Note that off-topic images do not belong to a set of pre-defined classes, making this problem different from the one where OIs belonged to a single class. To the best of our knowledge, the approach proposed in this paper is the first one in the literature to present this problem in the context of compressed sensing or group testing.
\end{enumerate}

In this paper, we combine the capabilities of neural networks and CS algorithms in a specific manner. Neural networks have been used in recent times for CS recovery \cite{Sun2020,Yin2022,Kulkarni2016} with excellent results. However, in this paper, we are using neural networks for either a noisy classification or a noisy outlier detection task, retaining the usage of classical CS or GT algorithms. 

\section{Group Testing Background}
\label{sec:background}
We first define some important terminology from the existing literature, which will be used in this paper.  In \textbf{binary group testing}, there are $n$ items, of which $k \ll n$ unknown items are defective.
Instead of individually testing each item, a pooled test (or group test) is performed on a pool/group of items to determine whether there exists at least one defective item in the pool, in which case the test is said to be `positive' (otherwise `negative'). The goal of group testing is to determine which $k$ items are defective with as few tests as possible. In \textbf{quantitative group testing}, pooled tests give the \emph{count} of the number of defective items in the pool being tested. In \textbf{non-adaptive group testing}, the number of tests and the pool memberships do not depend on the result of any test, and hence all tests can be performed in parallel. In \textbf{adaptive group testing}, the tests are divided into two or more rounds, such that the pool memberships for tests belonging to round $t_r+1$ depend on the test outcomes from rounds $[t_r] \triangleq \{1, \cdots, t_r\}$. \textbf{Dorfman Testing} is a popular $2$-round binary, adaptive algorithm \cite{Dorfman1943}, widely used in COVID-19 testing \cite{Tang2020}.
In round 1 of Dorfman testing, the $n$ items are randomly assigned to one out of $n/g$ groups, each of size $g$.
All items which are part of a negative pool are declared non-defective. All items which are part of a positive pool are then tested individually in round 2. The optimal pool size $g$ which minimizes the number of tests in the worst case is $\sqrt{n/k}$ and the number of tests is at most $2\sqrt{nk}$ \cite{Dorfman1943}.

A \textbf{pooling matrix} $\bPhi$ is a $m \times n$ binary matrix, where $\bPhi_{ij} =1$ indicates that item $j$ was tested as part of pool $i$ and $\bPhi_{ij} = 0$ otherwise. Let $\x$ be the unknown binary vector with $x_j = 1$ if item $j$ is defective, $0$ otherwise. Then the outcome of the pooled tests in binary group testing may be represented by the binary vector $\y$, such that $y_i = \bigvee_{i=1}^{n} \bPhi_{ij} \wedge x_j$ where $\vee$ and $\wedge$ denote the Boolean OR and AND operations respectively. The outcome of pooled tests in quantitative group testing may be represented by the integer-valued vector $\y = \bPhi \x$ obtained by multiplication of the matrix $\bPhi$ with the vector $\x$. In both binary and quantitative group testing, a \textbf{decoding algorithm} recovers $\x$ from known $\y$ and $\bPhi$. Group testing is termed \textbf{noisy} or \textbf{noiseless} depending on whether or not the test results $\boldsymbol{y}$ contain noise. 
\section{Main Approach}
\label{sec:main_approach}
\begin{algorithm}[t]
    \textbf{Input: } set of $n$ images $\mathcal{I}$, a trained \textsc{Qmpnn} or \Bmpnn (see Table~\ref{tab:abbreviations} and Sec.~\ref{subsec:pooled_nn}) parameterized by $\boldsymbol{\tilde \Theta}$ or $\boldsymbol \Theta$, $m\times n$ pooling matrix $\boldsymbol{\Phi}$, CS/non-adaptive GT decoding algorithm $\mathcal{F}$ \newline
    \textbf{Output: }binary vector $\x$ representing whether each of the images is OI or not 
    \begin{algorithmic}[1]
        \State If $\mathcal{F}$ is not a binary GT algorithm, run the \textsc{Qmpnn} on the $n$ images and obtain the predicted vector $\y$ of the number of OIs in each pool: 
        \newline $y_q \gets \underset{s\in \{0\dots r\}}{\arg\max }\: \qmpnn(\mathcal{I}, \boldsymbol{\Phi})(q, s)$  $\forall q \in \{1\dots m\}$,
        \State Otherwise, run the \textsc{Bmpnn} on the $n$ images and obtain the predicted vector $\bar{\y}$ indicating whether each pool contains an OI or not: \newline
        $\bar{y}_q \gets \underset{s\in \{0, 1\}}{\arg\max }\: \bmpnn(\mathcal{I}, \boldsymbol{\Phi})(q, s)$
        $\forall q \in \{1\dots m\}$, 
        \State Decode $\y$ or $\bar{\y}$ using the CS/non-adaptive GT algorithm $\mathcal{F}$ and obtain $\x \gets \mathcal{F}(\y, \boldsymbol{\Phi})$ or $\x \gets \mathcal{F}(\bar{\y}, \boldsymbol{\Phi})$
        \State \Return $\x$
    \end{algorithmic}
\caption{Pseudocode for Pooled Classification using CS/non-adaptive GT}  
\label{alg:inference}
\end{algorithm}
Fig.~\ref{fig:pooled_image_and_sfm} shows the main components of our approach for image moderation (see also Alg.~\ref{alg:inference}). Consider a set of $n$ images, $\mathcal{I} = \{I_1, \dots, I_n\}$, represented by a $n$-dimensional binary vector $\boldsymbol x$, with $x_i = 1$ if the $i^{\textrm{th}}$ image has objectionable content and $x_i = 0$ otherwise. The binary \textit{pooling matrix} $\boldsymbol \Phi$ of dimensions $m \times n$ specifies the $m$ image pools $\{P_1, \dots, P_m\}$, to be created from the images in $\mathcal{I}$. Each row of $\boldsymbol \Phi$ has sum equal to $r$, which means that each pool has $r$ images. The pooling matrix $\boldsymbol \Phi$ and the $n$ images in $\mathcal{I}$ are passed as input to a so-called \textit{quantitative matrix-pooled neural network} (\textsc{Qmpnn}, see Sec.~\ref{subsec:pooled_nn}) which is specifically trained to output the number of OIs in each pool in the form of a $m$-element vector $\boldsymbol{y}$. If the output of the \textsc{Qmpnn} is perfect, then clearly, for all $q \in \{1, \dots,  m\}$, $y_q = \sum_{I_i\in P_q} \mathbbm{1}(x_i = 1) = \sum_{i=1}^n {\boldsymbol \Phi_{qi}}x_i$, which gives us $\y = \boldsymbol{\Phi \x}$ just as in Sec.~\ref{sec:background}. Hence, recovery of the unknown vector of classifications $\x$ given the output $\y$ of \textsc{Qmpnn} (the vector of $m$ different quantitative group tests) and the pooling matrix $\boldsymbol \Phi$ can be framed as a CS problem. Such an approach is non-adaptive because the pool memberships for each image are decided beforehand, independent of the \textsc{Qmpnn} outputs. Furthermore, comparing with Eqn.~\ref{eq:forward}, we see that this CS problem has boolean constraints on each $x_i$, and with $y_q \in \{0, \dots, r\}$ for each $q$. 
In general the \textsc{Qmpnn} will not be perfect in reporting the number of OIs, hence the $\boldsymbol{y}$ vector predicted will be noisy. This is represented as:
\begin{equation}
\boldsymbol{y} = \mathfrak{N}(\boldsymbol{\Phi x}),
\label{eq:SI_forward}
\end{equation}
where $\mathfrak{N}(.)$ is a noise operator, which may not necessarily be additive or signal-independent unlike the case in Eqn.~\ref{eq:forward}.
However, we experimentally find that CS algorithms, which are specifically designed for signal-independent additive noise, are effective even in the case of noise in the outputs of the \textsc{Qmpnn}.
\subsection{Pooled Neural Network for Classification}
\label{subsec:pooled_nn}
The idea of a pooled neural network was first proposed in \cite{liang2021neural}.
We make two key changes to their design -- (1) incorporate the pooling scheme within the neural network which enables efficient non-adaptive (one-round)  testing, and (2) have quantitative outputs instead of binary, which enables CS decoding.
We discuss the significance of our architectural changes in detail in Sec.~\ref{subsec:differences}.
Below, we describe the pooled neural network architecture from \cite{liang2021neural} and then our network models. The rest of this section uses many abbreviations. For the reader's convenience, a complete list is presented in Table~\ref{tab:abbreviations}.

Consider a feed-forward deep neural network which takes as input a single image for classifying it as objectionable or not. 
We term such a network an \textit{individual neural network}, or \textsc{Inn}.
Furthermore, let this neural network have the property that it may be decomposed into $L$ blocks parameterized by the (mutually disjoint) sets of parameters 
$\theta_j, j \in \{1, \dots, L\}$, with $\boldsymbol \Theta \triangleq \bigcup_{j=1}^{L}\theta_j$, so that for any input image $I$,
\begin{equation}
\label{eq:inn}
    \inn(I) \triangleq \f{L} \circ \f{L-1} \circ \dots \circ \f{1}(I),
\end{equation}
where $\circ$ represents function composition, and where each block computes a function $\f{j}, j \in \{1, \dots, L\}$.
Furthermore, the $L^{\textrm{th}}$ block is a linear layer with bias, and has two outputs which are interpreted as the unnormalized log probability of the image being objectionable or not.
Below, we derive \textit{pooled neural networks} based on an \textsc{Inn} parameterized by $\boldsymbol \Theta$, and which take as input more than one image.

Consider the output of only the first $l < L$ blocks of the \textsc{Inn} on an input image $I$.
We call this the \textit{intermediate feature-map} (IFM) of image $I$, with    $\mathtt{IFM}_{\boldsymbol \Theta}(I) \triangleq \f{l} \circ \cdots \circ \f{1}(I)$. 
Consider a \textit{pool} created from $r$ images, given as $P \triangleq \{I_1, I_2, \dots , I_r\}$.
A \textit{superposed feature map} (SFM) of the pool of images $P$ is obtained by taking an entry-wise max over the feature-maps of individual images, i.e. $\mathtt{SFM}_{\boldsymbol \Theta}(P)(.) \triangleq \max_{I \in P} \mathtt{IFM}_{\boldsymbol \Theta}(I)(.)$ (known as `aggregated features' in \cite{liang2021neural}).
In a \textit{binary pooled neural network} (\textsc{Bpnn}), the SFM of a pool of images $P$ is processed by the remaining $L - l$ blocks of the \textsc{Inn}:
\begin{equation}
\label{eq:bpnn}
    \mathtt{BPNN}_{\boldsymbol \Theta}(P) \triangleq \f{L} \circ \f{L-1} \circ \dots \circ \f{l+1} \circ \mathtt{SFM}_{\boldsymbol \Theta}(P).
\end{equation}
The two outputs of a \textsc{Bpnn} are interpreted as the unnormalized log probability of the pool $P$ containing an image belonging to the OI class or not.
A \textsc{Bpnn} has the same number of parameters as the corresponding \textsc{Inn}.
This was introduced in \cite{liang2021neural}, where it is referred to as `Design 2'.

In this work, we introduce a \textit{quantitative pooled neural network} (\textsc{Qpnn}), which is the same as a \textsc{Bpnn} except that it has $r+1$ outputs instead of $2$. 
Due to this there are more parameters in the last linear layer (the $L^{\text{th}}$ block).
Let these parameters be denoted by $\tilde{\theta}_L$, with the $L^{\text{th}}$ block computing a function $f_{\tilde \theta_L}^{(L)}$, and 
$\mathtt{QPNN}_{\boldsymbol {\Tilde \Theta}}(P) \triangleq f_{\tilde \theta_{L}}^{(L)} \circ \dots \f{l+1} \circ \mathtt{SFM}_{\boldsymbol \Theta}(P)$,
with $\boldsymbol{ \tilde \Theta} \triangleq \bigcup_{j=1}^{L-1}\theta_j \cup \tilde{\theta}_L.$
The $r+1$ outputs of a \textsc{Qpnn} are interpreted as the unnormalized log probabilities of the pool $P$ containing $0$ through $r$ instances of images which belong to the OI class.

Finally, we introduce the \textit{quantitative matrix-pooled neural network} (\textsc{Qmpnn}), which takes as input a set of $n$ images $\mathcal{I} \triangleq \{I_1, \dots, I_n\}$, a specification of $m$ pools containing $r$ out of $n$ images in each pool via binary matrix $\boldsymbol{\Phi}$, and outputs $r+1$ real numbers for each of the $m$ pools, in a single forward pass.
That is,
\begin{align}
    \mathtt{QMPNN}_{\boldsymbol{\tilde \Theta}}(\mathcal{I}, \boldsymbol \Phi) &\triangleq    (\mathtt{QPNN}_{\boldsymbol {\Tilde \Theta}}(P_1),
    \dots,
    \mathtt{QPNN}_{\boldsymbol {\Tilde \Theta}}(P_m) ) \\
    \text{ where } P_q &\triangleq \{I_i | \Phi_{qi} = 1, I_i\in \mathcal{I}\} \text{ }\forall q \in \{1,\dots, m\}. \nonumber
\end{align}
For a pool $q$, the $s^\text{th}$ output ($s \in \{0, \dots, r\}$) is interpreted as the unnormalized log probability of the pool $q$ containing $s$ images belonging to the OI class.
Notably, the IFM for each of the $n$ images is computed only once by the \Qmpnn, and is re-used for each pool that it takes part in.
A \textit{binary matrix-pooled neural network} (\textsc{Bmpnn}) is exactly the same as a \Qmpnn except that each pool has only two outputs. 
We note that while the first $l$ blocks of the neural network are run for each of the $n$ images, the remaining $L-l$ blocks are only run for each of $m$ SFMs.
However, if an individual neural network (\textsc{Inn}) is run on the $n$ images, then both the first $l$ blocks and the last $L - l$ blocks will be run $n$ times.
Hence, since $m < n$, the \Qmpnn (and the \Bmpnn) requires significantly less computation than running the \textsc{Inn} on the $n$ images, which we verify empirically in Sec.~\ref{sec:exp}.

\textbf{Training:}
Let $\mathcal{D}$ be a dataset of images labelled as OI or non-OI. A \textit{pooled dataset} $\mathcal{D}_{\text{pooled}}$ is obtained from $\mathcal{D}$.
Each entry of $\mathcal{D}_{\text{pooled}}$ is a pool of $r$ images, and has a label in $\{0,\dots, r\}$, equal to the number of images in that pool which belong to the OI class.
Details of creation of the pooled dataset are given in Sec.~\ref{sec:exp}.
The parameters $\boldsymbol{\tilde \Theta}$ of a \textsc{Qpnn} are trained via supervised learning on $\mathcal{D}_{\text{pooled}}$ using the cross-entropy loss function, $
    L(\boldsymbol{\tilde \Theta}) = \underset{(P, c) \sim \mathcal{D}_{\text{pooled}}}{E} [- \log \sigma(\qpnn(P))_c],
$
where $\sigma(.)$ is the softmax function, given by $\sigma(\boldsymbol z)_s \triangleq \frac{e^{z_s}}{\sum_{t=0}^r e^{z_t}}$. The parameters $\Theta$ of a \Bpnn are trained similarly, except that each training pool has a label in $\{0, 1\}$, with $1$ indicating pools which contain at least one OI.

\textbf{Inference:}
The trained parameters $\boldsymbol{\tilde \Theta}$, and a pooling matrix $\bPhi$ are then used to instantiate a \Qmpnn.
The entries of the vector $\boldsymbol y$, containing pool-level predictions of the count of OIs in each pool, are obtained from a \textsc{Qmpnn} as:
\begin{equation}
\label{eq:qmpnn_output}
    y_q = \underset{s \in \{0,1\dots, r\}}{\arg\max}\text{ } \mathtt{QMPNN}_{\boldsymbol{\tilde \Theta}}(\mathcal{I}, \boldsymbol \Phi) (q, s) \text{ }\forall q \in \{1,\dots, m\}.
\end{equation}
In Sec.~\ref{subsec:algorithms}, we present CS algorithms to decode this vector $\boldsymbol y$ to recover the vector $\boldsymbol{x}$ for classification of each image in $\mathcal{I}$.

Similarly, the trained parameters $\boldsymbol{\Theta}$ of a $\Bpnn$ and a pooling matrix $\boldsymbol{\Phi}$ can be used to instantiate a \Bmpnn. 
The entries of the binary vector $\bar{\boldsymbol{y}}$ containing pool-level predictions of whether each pool contains an OI or not can be obtained from the \textsc{Bmpnn} as:
\begin{equation}
\label{eq:bmpnn_output}
    \bar{y}_q = \underset{s \in \{0,1\}}{\arg\max}\text{ } \mathtt{BMPNN}_{\boldsymbol{\Theta}}(\mathcal{I}, \boldsymbol \Phi) (q, s) \text{ }\forall q \in \{1,\dots, m\}.
\end{equation}
This can be decoded by the non-adaptive binary group testing decoding algorithms presented in Sec.~\ref{subsec:algorithms}.

\subsection{CS/GT Decoding Algorithms}
\label{subsec:algorithms}
Given Eqn.~\ref{eq:SI_forward}, the main aim is to recover $\boldsymbol{x}$ from $\boldsymbol{y}$ (the prediction of the \textsc{Qmpnn}, Eqn.~\ref{eq:qmpnn_output}) and $\boldsymbol{\Phi}$. For this, we propose to employ the following two algorithms:
\begin{enumerate}
    \item Mixed Integer Programming Method (\textsc{Mip}):
    We minimize the objective in Eqn. \ref{eq:lasso} with boolean constraints on entries of $\x$, i.e. $\forall i\in [n],      x_i \in \{0, 1\}$.
    We use the CVXPY \cite{diamond2016cvxpy} \cite{agrawal2018rewriting} package with the Gurobi \cite{gurobi} solver, which uses a branch and bound method for optimization \cite{milp}.
    \item Constrained LASSO (\textsc{CLasso}): This is a variant of the \textsc{Lasso} from Eqn.~\ref{eq:lasso}, with the values in $\boldsymbol{x}$ constrained to lie in $[0,1]$.
    The relaxation from $\{0,1\}$ to $[0,1]$ is done for computational efficiency.
    The final estimate regarding whether each of the images is an OI is obtained by thresholding $\boldsymbol{\hat{x}}$ 
    , i.e. $\forall i \in [n]$, $I_i$ is considered to be an OI if $\hat{x}_i > \tau$, where $\tau$ is the threshold determined on a validation set.
\end{enumerate}
The hyperparameters in our algorithms ($\lambda$ for \classo and \milp, and $\tau$ for \classo) are chosen via grid search so as to maximize the product of specificity and sensitivity (both defined in Sec.~\ref{sec:exp}) on a representative validation set of images of known classes.

We also compare with some popular algorithms the from binary group testing literature, which act as a baseline.
For the non-adaptive methods, the binary vector $\bar{\boldsymbol y}$  containing predictions of a \Bmpnn (from Eqn.~\ref{eq:bmpnn_output}) is provided to the decoding algorithms. The algorithms are:
\begin{enumerate}
    \item Dorfman Testing: See Sec.~\ref{sec:background}.
    This is referred to as Two-Round testing in \cite[Algorithm 1]{liang2021neural}. In the first stage, a binary pooled neural network (\Bpnn) is used to classify a pool of images as containing an objectionable image or not.
    If the \Bpnn predicts that the pool contains an objectionable image, then each image is tested individually in a second round using an individual neural network (\textsc{Inn}).
    Otherwise, all images in the pool are declared as not being objectionable, without testing them individually.
    Because of a second round of testing, Dorfman Testing is inherently resistant to a pool being falsely declared as positive, but not to a pool being falsely declared as negative.
    \item Combinatorial Orthogonal Matching Pursuit (\textsc{Comp}) \cite{chan2011non}:
    This is a simple decoding algorithm for noiseless non-adaptive binary group testing, wherein any image which takes part in at least one pool which tests negative (i.e., no OIs in the pool) is declared as a non-OI, and the remaining image are declared as OIs.
    The One-Round Testing method in \cite{liang2021neural} uses \comp decoding, albeit with a different pooling matrix than ours.
    \item Noisy \comp (\ncomp) \cite{chan2011non}: a noise-resistant version of \comp. An image is declared to be OI if it takes part in strictly greater than $t$ positive pools (or equivalently in strictly less than $c-t$ negative pools, where $c$ is the number pools that each image takes part in).
\end{enumerate}
A good pooling matrix $\boldsymbol{\Phi}$ is crucial for the performance of these algorithms.
We choose binary matrices with equal number of ones in each column, equal number of ones in each row, with the dot product of each pair of columns and each pair of rows at most $1$.
More details about this are provided in Appendix~\ref{sec:pooling_matrix}.

\subsection{Pooled Deep Outlier Detection}
\label{subsec:outlierdetection}

\begin{algorithm}[t]
    \textbf{Inputs: } set of $n$ images $\mathcal{I}$,
    trained \textsc{Qmpnn} parameterized by $\boldsymbol{\tilde \Theta}$,
    $m\times n$ pooling matrix $\boldsymbol{\Phi}$,\newline
    \hspace*{\algorithmicindent} GMM $G$ fit to feature vectors of training pools with no off-topic images,\newline
    \hspace*{\algorithmicindent} histogram of training pool anomaly scores $H_G$, \newline
    \hspace*{\algorithmicindent} CS/non-adaptive GT decoding algorithm $\mathcal{F}$ \newline
    \textbf{Output: }binary vector $\x$ representing whether each of the images are off-topic or on-topic 
    \begin{algorithmic}[1]
        \State Run the \textsc{Qmpnn} with pooling matrix $\boldsymbol{\Phi}$ on the $n$ images and obtain $m$ pools $\{P_1,\dots,P_m\}$ and their feature vectors $\{\phi(P_1),\dots, \phi(P_m)\}$
        \State Use the GMM $G$ to obtain the vector of anomaly scores for the $m$ pools,
        \newline $S := \{\nlpd_G(\phi(P_1)),\dots,\nlpd_G(\phi(P_m))\}$
        \State Find the bin indices of the anomaly scores in $S$ in the histogram $H_G$ and use Eqn.~\ref{eq:label_pool} to obtain each entry of the predicted vector $\boldsymbol y$ of the number of off-topic images in the pools
        \State If $\mathcal{F}$ is a binary GT algorithm, binarize $\boldsymbol y$ by setting each non-zero entry to 1
        \State Decode $\y$ using the CS/non-adaptive GT algorithm and obtain $\x \gets \mathcal{F}(\y, \boldsymbol{\Phi})$
        \State \Return $\x$
    \end{algorithmic}
\caption{Pseudocode for Pooled Outlier Detection using CS/non-adaptive GT} 
\label{alg:ood_inference}
\end{algorithm}

Here, we consider the case where the set of allowed images belong to one underlying class, and any image not belonging to that class is considered `off-topic'.
Such a situation arises in the moderation of topical communities on online forums such as Reddit (e.g. see \cite{jhaver2019human}).
For example, an image of a baseball game is off-topic on a forum meant for sharing of tennis-related images.
As the off-topic images could belong to an unbounded variety of classes, instances of all or some of which may not be available for `training', \textit{classification} based approaches (such as the \Qmpnn from Sec.~\ref{subsec:pooled_nn}, and \cite{liang2021neural}) may not be directly applicable, and instead, \textit{deep outlier detection} based methods are more suitable. 

Recent deep outlier detection methods \cite{lee2018simple, sehwag2020ssd} approximate a class of interest, represented by suitable feature vectors in $\mathbb{R}^d$, using a high dimensional Gaussian distribution characterized by a mean vector $\boldsymbol{\mu} \in \mathbb{R}^d$ and a covariance matrix $\boldsymbol{\Sigma} \in \mathbb{R}^{d \times d}$.
Then the Mahalanobis distance between the feature vectors of a test image and the mean vector $\boldsymbol{\mu}$ is computed.
This Mahalanobis distance acts as the anomaly score to perform outlier detection, i.e. test feature vectors with a Mahalanobis distance greater than some threshold (say $\bar{\tau}_{md}$) are considered outliers. The feature vectors can be represented by the outputs of a suitably trained neural network. One may think of using such an approach with a pooled neural network to detect pools containing outlier images. However a single Gaussian distribution is often inadequate to represent a sufficiently diverse class, and hence we resort to using a Gaussian Mixture Model (GMM).
We put forward a novel approach which uses a pooled neural network  combined with a GMM, to detect anomalous pools, i.e., pools which contain at least one off-topic image, which we term \emph{pooled deep outlier detection}.
This enables us to use binary group testing methods from Sec.~\ref{subsec:algorithms} - such as \comp, \ncomp and Dorfman Testing - for pooled outlier detection. Furthermore, we also detect the number of outlier images in a pool, thus enabling CS methods to be used for pooled outlier detection.
To the best of our knowledge, this is the first such work in the literature, which enables group testing and compressed sensing methods to be used for outlier detection. 

We consider the setting where for training we have available a set of images which are either on-topic or are off-topic images from known classes, but at test time the off-topic images may be from unknown classes. We train a \textsc{Qpnn} using the method in Sec.~\ref{subsec:pooled_nn}, with the label $L(P)$ for any training pool $P$ set to the number of off-topic images in it.
The output of the last-but-one layer of the \textsc{Qpnn} (i.e. the $(L-1)^{\text{th}}$ layer -- see Sec.~\ref{subsec:pooled_nn}) is considered to be the feature vector $\phi(P)$ for the pool of images $P$ which is input to it.
First, we create $M$ pools from the on-topic training data images (i.e. these pools have $0$ off-topic images), and fit a GMM $G$ with some $K$ clusters to the feature vector $\phi(.)$ of all these pools, using the well-known expectation-maximization (EM) algorithm.
The optimal value of $K$ is chosen via cross-validation, by selecting the $K$ with the maximum likelihood given a held-out set of pools containing only on-topic images.

\begin{algorithm}[t]
    \textbf{Inputs: } Pool $P := \{I_1,\dots,I_r\}$ of $r$ images,
    trained \textsc{Qpnn} parameterized by $\boldsymbol{\tilde \Theta_1}$,
    trained \textsc{Inn} parameterized by $\boldsymbol{\Theta_2}$,\newline
    \hspace*{\algorithmicindent} GMM $G_1$ fit to feature vectors of training image pools with no off-topic images,\newline
    \hspace*{\algorithmicindent} Histogram $H_{G_1}$ of anomaly scores of training image pools, 
    \newline
    \hspace*{\algorithmicindent} GMM $G_2$ fit to feature vectors of on-topic training images, \newline 
    \hspace*{\algorithmicindent} Histogram $H_{G_2}$ of anomaly scores of training images,
    \newline
    \textbf{Output: }binary vector $\x$ representing whether each of the images are off-topic or on-topic 
    \begin{algorithmic}[1]
        \State Run the \textsc{Qpnn} on pool $P$ and obtain its feature vector $\phi(P)$
        \State Use the GMM $G_1$ to obtain the anomaly score for the pool, $\nlpd_{G_1}(\phi(P))$,
        \State Find the bin index of the anomaly score $\nlpd_{G_1}(\phi(P))$ in the histogram $H_{G_1}$ and use Eqn.~\ref{eq:label_pool} to obtain the number of off-topic images in pool $P$, $L(P)$
        \State If $L(P)$ is 0, output is vector of all zeros, i.e., $\boldsymbol x \gets{\boldsymbol 0}$ and \Return $\x$
        \State Otherwise, run the \textsc{Inn} on each image $\{I_1,\dots,I_r\}$ to obtain $r$ image feature vectors, $\{\phi(I_1),\dots,\phi(I_r)\}$
        \State Obtain the vector of anomaly scores of the images using the GMM $G_2$, $S := \{\nlpd_{G_2}(\phi(I_1),\dots,\nlpd_{G_2}(\phi(I_r)\}$
        \State Find the bin indices of the anomaly scores in $S$ in the histogram $H_{G_2}$ and obtain the set of labels of each image $\{L(I_1),\dots,L(I_r)\}$, similar to Eqn.~\ref{eq:label_pool}
        \State Set $\x \gets (L(I_1),\dots,L(I_r))$ and \Return $\x$
    \end{algorithmic}
\caption{Pseudocode for Dorfman Pooled Outlier Detection} 
\label{alg:ood_dorfman}
\end{algorithm}

For any pool $P$, we use the negative log probability density of its feature vector $\nlpd_G(\phi(P))$ (negative log probability density) under the GMM $G$ as an anomaly score, given by:
\begin{equation}
 \nlpd_G(\phi(P)) := -\log\left(\sum_{j=1}^{K} p_j \mathcal{N}(\phi(P)|\boldsymbol{\mu_j},\boldsymbol{\Sigma_j})\right),
 \label{eq:NLL}
\end{equation}
where $\{\boldsymbol{\mu_1}, \boldsymbol{\mu_2}, \cdots, \boldsymbol{\mu_K}\}$ are the $K$ mean vectors,
$\{\boldsymbol{\Sigma_1}, \boldsymbol{\Sigma_2},\cdots,\boldsymbol{\Sigma_K}\}$ are the $K$ covariance matrices
and $\{p_1, p_2, \cdots, p_K\}$ are the $K$ membership probabilities of the GMM $G$,
and for any vector $\mathbf{z}$, $\mathcal{N}(\mathbf{z}|\boldsymbol{\mu_j},\boldsymbol{\Sigma_j})$ is its probability density given the multivariate normal distribution with mean $\boldsymbol{\mu_j}$ and covariance matrix $\boldsymbol{\Sigma_j}$. 

It is intuitive to imagine that $\nlpd_G(\phi(P))$ values would generally  be larger for pools containing a larger number of outlier images ($L(P)$), and we should be able to predict $L(P)$ given $\nlpd_G(\phi(P))$. The main idea is to create a histogram of anomaly scores of pools with different number of off-topic images in them, and assign to a new pool the label which is most common in its bin. This is accomplished as follows.
From the \emph{validation} set of images, some $N$ pools $\{P_i\}_{i=1}^N$ are created, each containing $r$ images.
Since off-topic images are expected to be rare at test time, we only consider pools which contain upto some $t < r$ off-topic images.
Thus $L(P) \in \{0,1,\dots, t\}$.
For each pool $P_i$, the value $\nlpd_G(\phi(P_i))$ is computed.
These NLPD values are divided into some $Q$ bins/intervals and an anomaly score histogram $H_G$ is created in the following manner:
\begin{equation}
    \forall j \in \{1,2,\cdots,Q\}, \forall l \in \{0,1,\cdots,t\},  H_G(j,l) = \dfrac{\# \textrm{pools in bin } j \textrm{ containing } l \textrm{ outlier images}}{N}.
\label{eq:histjl}    
\end{equation}
That is, $H_G(j,l)$ represents the fraction of pools from $\{P_i\}_{i=1}^N$ which contain $l$ outlier images and whose NLPD value falls into the $j^{\text{th}}$ bin.
This histogram $H_G$ is created at the time of training.

Each  non-empty bin $j$ is assigned a label $\mathcal{L}(j)$ equal to $\textrm{argmax}_{l \in \{0,1,\cdots,t\}} H_G(j,l)$, i.e., the label with the most number of pools in that bin. 
   An empty bin is assigned the label of its nearest non-empty bin.
   If there is more than one nearest non-empty bin, then the bin is assigned the label of the bin with the larger index.
   At test time, given the feature vectors $\phi(P_{\text{test}})$ of a pool $P_{\text{test}}$ created from some $r$ test images, first its anomaly score $\nlpd_G(\phi(P_{\text{test}}))$ is computed using the GMM.
   If the anomaly score lies in bin $j_{\text{test}}$, then $P_{\text{test}}$ is assigned the label $\mathcal{L}(j_{\text{test}})$.
   If the anomaly score is to the right of all bins (i.e. greater than the right index of the rightmost bin), then it is assigned the label $t$.
   If it is to the left of all bins (i.e. less than the left index of the leftmost bin), then it is assigned the label $0$.
   That is, 
   \begin{equation} \label{eq:label_pool}
       L(P_{\text{test}}) = 
       \begin{cases}
       \mathcal{L}(j_{\text{test}}) & \text{if } \nlpd_G(\phi(P_{\text{test}})) \in \text{ bin } j_{\text{test}} \\
       0 & \text{if } \nlpd_G(\phi(P_{\text{test}})) < s_{\text{min}} \\
       t & \text{if } \nlpd_G(\phi(P_{\text{test}})) > s_{\text{max}} \\
       \end{cases},
   \end{equation}
    where $s_{\text{min}}$ and $s_{\text{max}}$ are respectively the smallest and the largest of the $N$ NLPD values used to create the histogram $H$.

At the time of deployment, $m$ pools are created from $n$ images to be tested for off-topic content using a \Qmpnn and a pooling matrix $\Phi$, and feature vectors of the $m$ pools are obtained.
For each pool, we determine the number of off-topic images it contains using the GMM and histogram-based method just described.
Given these $m$ numbers and the pooling matrix $\Phi$, we predict whether or not each of the $n$ images is off-topic using the CS and non-adaptive GT decoding algorithms described in Sec.~\ref{subsec:algorithms}. Our method is summarized in Algorithm~\ref{alg:ood_inference}.

We also perform Dorfman Testing for outlier detection.
Feature vectors of a pool of $r$ images are obtained by running a \Qpnn, and the GMM and histogram-based method is used to predict the number of off-topic images in the pool.
If the pool is predicted to contain at least one off-topic image, then each image is individually tested for being off-topic using the same method, but using an individual neural network (\textsc{Inn}).
That is, the features of each image are obtained using the \textsc{Inn}, and a previously trained GMM and histogram (of validation \emph{image} anomaly scores) are used to label the images as off-topic or not.
This is summarized in Algorithm.~\ref{alg:ood_dorfman}.

\subsection{Comparison with Related Work} 
\label{subsec:differences}
There exists very little literature on the combination of GT algorithms with neural networks. The sole published work on this topic (to our knowledge) can be found in \cite{liang2021neural}.
The neural networks in \cite{liang2021neural} use only binary outputs, while the neural networks proposed in our methods have quantitative outputs, and are trained to predict the \emph{number of OIs} in a given SFM.
This enables usage of CS decoding algorithms.
CS  methods have better recovery guarantees than non-adaptive binary GT (see Appendix~\ref{sec:recovery_guarantees}).
Non-adaptive pooling using a $m\times n$ pooling matrix is implemented in \cite{liang2021neural} by separately running a \textsc{Bpnn} for each pool.
In their method, the intermediate feature map (IFM) for an image needs to be re-computed for each pool that it takes part in, making it computationally inefficient.
Due to this, the scheme in \cite{liang2021neural} is limited to using pooling matrices in which each image takes part in at most two pools.
Such matrices can be at most $1$-disjunct (see Appendix~\ref{sec:properties_sensingmatrices} for the definition), and hence guarantee of exact recovery via \comp is for only $1$ OI per matrix (see \cite[Prop. 2.1]{erdos1985families} and Appendix~\ref{sec:disjunctness_upper_bound} for an explanation).
Thus their scheme is applicable for only very low prevalence rates ($p$) of OIs.
In our implementation, an IFM for each image is computed only once, and is re-used for each pool that it takes part in.
This enables usage of pooling matrices with $3$ or more entries per column.
This means that we can use matrices with high `disjunctness' for group testing decoding \cite{Mazumdar2012}
or those obeying restricted isometry property (RIP) of high order \cite{berinde2008combining} for compressed sensing recovery
(see Appendix~\ref{sec:properties_sensingmatrices} for definitions of disjunctness and RIP).
Due to this, our method is applicable for both low or high $p$ values.
Furthermore, there are no guarantees for binary group testing for recovery from noisy pool observations using matrices which have only two $1$ entries per column.
This is because if one test involving an item is incorrect but the other one is correct, there is no way to determine which of them is the incorrect one.
Noise-tolerant recovery algorithms for binary GT exists for suitable matrices with $3$ or more $1$ entries per column.
For example, \ncomp with a pooling matrix which has $3$ tests per item and has the properties described in Appendix~\ref{sec:pooling_matrix}, when testing in the presence of upto one defective item, can recover from an error in one test of that item by declaring it defective if two of the tests are positive, or non-defective if two of its tests are negative.
Moreover, higher number of $1$ entries per column implies better noise tolerance for CS algorithms for the matrices described in Appendix~\ref{sec:pooling_matrix}.
See Appendix~\ref{sec:binary_matrices_noise_tolerance} for details.
 
Over and above these differences, we also extend binary group testing and compressed sensing to deep outlier detection as earlier described in Sec.~\ref{subsec:outlierdetection}, whereas the approach of \cite{liang2021neural} is limited to the classification setting.
In the case of OIs with a single class, our test datasets also contain a much larger number of objectionable images (see details in Sec.~\ref{sec:exp}), and these are mixed with non-OIs to form datasets of size 1M or 100K for a wide range of prevalence rates $p$, so that each OI gets tested many times in combination with other OIs and different non-OIs.
However in \cite{liang2021neural}, prediction measures are presented only for a single dataset for the case when $p=0.001$, with the number of unique OIs being only $50$ and each OI being tested only once. 
When a single OI participates in multiple pools, there is an inherent noise resilience brought in.
This is important as the neural network may not produce perfect classification.

\section{Experiments}
\label{sec:exp}
We present an extensive set of experimental results, first focusing on the classification case, and then on the outlier detection case.

\subsection{Image Moderation using Pooled Classification: Single Class for Objectionable Images}
\label{subsec:single_class}
Here the objectionable images belong to a single underlying class, and the goal is to classify images as objectionable or not objectionable.

\paragraph{\textbf{Tasks:}}
We use the following two tasks for evaluation:
(\textit{i}) firearms classification using the Internet Movie Firearms Database (IMFDB) \cite{imfdb}, a popular dataset used for firearms classification \cite{Bhatti2021,Debnath2021}; and
(\textit{ii}) Knife classification using the Knives  dataset \cite{Grega2016}, a dataset of CCTV images, labelled with the presence or absence of a handheld knife.
In (\textit{i}), the firearm images from IMFDB are mixed with non-firearm images from ImageNet-1K  \cite{imagenet2012} as IMFDB contains only firearm images. The firearms and the knife images are respectively considered as objectionable images (OIs) in the two tasks. Here below, we provide details of the number of images in the training/validation/test splits in each dataset.

\paragraph{\textbf{Individual Dataset Splits:}}
For the firearm classification task, we use firearm images from IMFDB \cite{imfdb} and Imagenet-1K \cite{imagenet2012} as the OI class, with \num{9458} and \num{3617} (total \num{13075}) images in the training set from the two datasets respectively. Additionally, 667 firearam images from IMFDB were used in the validation set, and 931 images were used in the test set. The validation and test images were manually cleaned to remove images which did not contain a firearm, or images in which the firearm was not the primary object of interest. For the non-OI images, we used images from 976 classes in ImageNet-1K \cite{imagenet2012}, which did not belong to a firearm class -- \num{1200764} in the training set, \num{50000} in the validation set, and \num{48800} in the test set (leaving out weapon images from ImageNet-1K such as `tank', `holster', `cannon', etc., following \cite{liang2021neural}). During training of the \textsc{Inn}, the classes were balanced by selecting all \num{13075} OIs and the same number of random non-OIs for training at each epoch. Classes were balanced in the validation split as well, by using all \num{50000} non-OIs and randomly sampling with replacement \num{50000} OIs.
\begin{figure}
\centering
    \includegraphics[width=0.1\columnwidth]{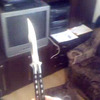}
    \includegraphics[width=0.1\columnwidth]{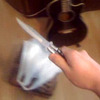}
    \includegraphics[width=0.1\columnwidth]{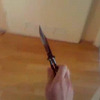}
    \\
    \includegraphics[width=0.1\columnwidth]{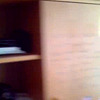}
    \includegraphics[width=0.1\columnwidth]{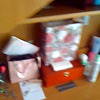}
    \includegraphics[width=0.1\columnwidth]{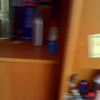}
    \centering
    \caption{Data Samples from Knives dataset \cite{Grega2016}. Top Row: Images containing a knife, Bottom Row: Images not containing a knife.} 
    \label{fig:knives}
\end{figure}

The Knives dataset \cite{Grega2016} used for the knife classification task  consists of 12,899 images of which 3559 are knife images and 9340 are non-knife images.
Some images were taken indoors, while some were taken through a car window in the street.
We randomly selected 2159 knife and 6140 non-knife images for training, 700 knife and 1600 non-knife images for validation, and 700 knife and 1600 non-knife images for testing.
During training of the individual neural network, classes were balanced by sampling (without replacement) 2159 random non-knife images at each epoch. The validation set was also balanced by sampling (with replacement) 1600 random knife images.
See Fig.~\ref{fig:pooled_image_and_sfm} for examples of images in IMFDB, and Fig.~\ref{fig:knives} for examples of images from the Knives dataset.
\paragraph{\textbf{Pooled Dataset Splits:}}
For firearms classification, during training of the \textsc{Qpnn}, at each epoch, \num{6248} pools of size $r=8$ were created, each pool containing $k$ randomly selected OIs and $8-k$ randomly selected non-OIs, for each $k$ in $\{0\dots8\}$.
Of these, $40\%$ pools contained no OIs, $24\%$ contained 1 OI, $12\%$ contained 2 OIs, $6\%$ each containing 3 and 4 OIs, and $3\%$ each contained 5 through 8 OIs. Such a distribution was chosen because OIs are assumed to have low prevalence rate in the test dataset. Hence greater accuracy is desired on pools with low number of OIs -- however, enough instances of pools with high number of OIs must be seen during training for good accuracy.
The pooled validation set used for selection of neural network weights contained \num{63325} pools of size $8$, created with the same method as the pooled training set. For testing, we created $9$ mixtures of the test data split of the individual dataset,
each containing $N=1 \text{ Million}$ images, with
each mixture containing $pN$ OIs, with the prevalence rate $p \in \{0.001, 0.002, 0.005, 0.01, 0.02, 0.03, 0.04, 0.05, 0.1\}$. 

For knife classification, 12496 pools of size 8 were created for training following the same distribution of OIs as for the firearms classification case.
The pooled validation set used for selection of neural network weights contained 2873 pools of size 8, created with the same method as the pooled training set.
The pooled test dataset consisted of $9$ mixtures of 100K images, created with the same method as for the firearms classification task.

\paragraph{\textbf{Data Transformations:}}
Images were padded with zeros to equalize their height and width, and resized to $224\times 224$, during both training and testing. We also added a random horizontal flip during training. For the pooled datasets, a random rotation rotation between $[\ang{-30}, \ang{30}]$ was applied to each training image.
This rotation was also applied to test data for the knife classification task.

\paragraph{\textbf{Training:}}
We train all our networks for 90 epochs using stochastic gradient descent with learning rate $0.001$ and momentum $0.9$, using cross-entropy loss and weight decay of $0.0001$. Model weights are checkpointed after every epoch. The best model weights are selected from these $90$ checkpoints using classification accuracy on the validation set.
For the firearms classification task, a weighted accuracy is used for selection of the best checkpoint of the pooled neural networks, viz $\mathtt{Acc}_{\textrm{weighted}} = \sum_{k=0}^8 \mathtt{Binom}(8, p, k) \mathtt{Acc}_k$, where $\mathtt{Acc}_k$ is the classification accuracy on pools with $k$ OIs in them, and $\mathtt{Binom}(8, p, k)$ is the probability of having $k$ OIs in a pool of size $8$ if the prevalence rate of OIs is $p$.
This accounts for the fact that accuracy on pools with higher number of OIs is more important for high values of $p$, while pools with $0$ or $1$ OI are more important for low values of $p$.

\begin{figure}
    \centering
    \includegraphics[width=0.5\textwidth]{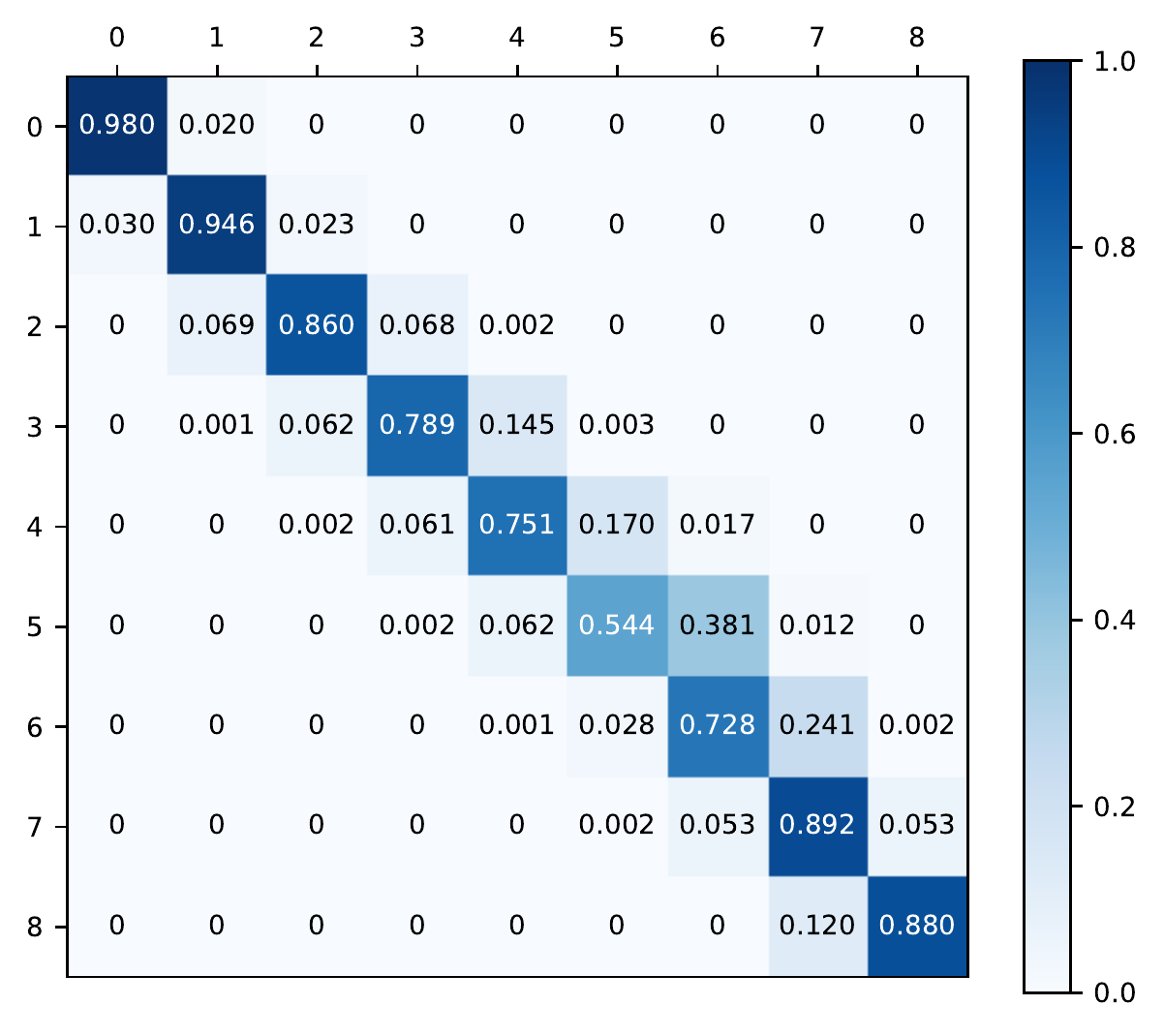}
    \caption{\textsc{Qmpnn} Confusion Matrix on IMFDB Pooled Validation Data: true \#OIs (rows) versus predicted \#OIs (columns)}
    \label{fig:confusion_matrix}
\end{figure}

\paragraph{\textbf{Confusion Matrix:}}
As can be seen in Fig. \ref{fig:confusion_matrix}, if $g$ is the ground truth number of OIs, the predictions of the \textsc{Qmpnn} lie in the interval $[\textrm{max}(0,g-1),\textrm{min}(g+1,r)]$ with high probability. This is presented as a confusion matrix of size $(r+1) \times (r+1)$ of the ground truth number of OIs in a pool versus the predicted number. In Fig.~\ref{fig:confusion_matrix}, we have $r=8$.
\paragraph{\textbf{Neural Network Implementation:}}
We use the official PyTorch implementation of ResNeXt-101 ($32 \times 8d$) \cite{xie2017resnext} \cite{pytorchresnext},
a highly competitive architecture which has produced compelling results in many image classification tasks,
with weights pre-trained on ImageNet-1K. The last linear layer of this network has $1000$ outputs by default.
Recall the discussion in Sec.~\ref{subsec:pooled_nn} regarding the architectures of the \textsc{Inn}, \Qmpnn, \Bmpnn, \Qpnn, and \Bpnn. For \textsc{Qpnn}/\textsc{Qmpnn} with pool size $r = 8$, we replace the last linear layer to have only $9$ outputs.
For the \textsc{Inn} or \textsc{Bpnn}/\textsc{Bmpnn}, we replace the last linear layer to have only two outputs.
Each image is passed through first three stages of the ResNeXt-101 to create IFMs, which are then combined to create SFMs for the pools in the \Qmpnn/\Qpnn/\Bmpnn/\Bpnn.
The remaining two stages of the ResNeXt-101 process only the SFMs in \Qmpnn/\Qpnn/\Bmpnn/\Bpnn.
This configuration is exactly the same as Design 2 in \cite[Sec. II.A]{liang2021neural}.
In initial experiments, we also tried using Design 3 of \cite{liang2021neural}, but the accuracy of this configuration was not good on the IMFDB dataset.
The networks are trained using the method specified in Sec.~\ref{subsec:pooled_nn}.
Details of training the individual and pooled neural networks have been given earlier.
\paragraph{\textbf{Pooling Matrix:}}
We use a $50\times 100$ balanced pooling matrix $\bPhi$ with the construction as described in Appendix \ref{sec:pooling_matrix}, with $r=8$ ones per row and $c=4$ ones per column.
That is, $50$ SFMs are created from $100$ IFMs, with each SFM being created from $8$ different IFMs, and each IFM contributing to $4$ different SFMs.

\paragraph{\textbf{Prediction:}}
We randomly sampled $pN$ OIs and $(1-p)N$ non-OIs from the test split and shuffled these $N$ images to create our test datasets.
Here, $p \in \{0.001, 0.002, 0.005, 0.01, 0.02, 0.03, 0.04, 0.05, 0.1\}$ is the prevalence rate of OIs, and $N$ is 1M for IMFDB and 100K for Knives.
Each test set is divided into chunks of size $100$ each and passed to the \textsc{Qmpnn} to retrieve the prediction vector $\y$ (of length $50$ ) for each chunk.
As can be seen in Fig~\ref{fig:confusion_matrix}, if $g$ is the ground truth number of OIs, the predictions lie in the interval $[\textrm{max}(0,g-1),\textrm{min}(g+1,r)]$ with high probability.
These pool-level predictions are decoded using the algorithms \textsc{CLasso} and  \textsc{Mip}.
We compare them with the baselines of \textsc{Comp}, Dorfman Testing with pool-size 8
(\dorf-8, same as Design 2 + Algorithm 1 of \cite{liang2021neural}), \textsc{NComp} with $t=2$ (\ncomp-2) and Individual Testing (\individual) on images of the same size. The numerical comparison was based on the following two performance measures:
\begin{enumerate}
    \item \textit{Sensitivity} (also called \textit{Recall} or True Positive rate) $\triangleq \#$correctly detected OIs / $\#$actual OIs
    \item \textit{Specificity} (also called True Negative Rate) $\triangleq \#$correctly detected non-OIs / $\#$actual non-OIs. 
\end{enumerate}
For both the measures, larger values indicate better performance.

\paragraph{\textbf{Hyperparameter Selection:}}
We created mixtures of images from the validation split of the same sizes and the same prevalence rates as for the test split. 
For a given prevalence rate $p$, grid search was performed on the corresponding validation split mixture with the same value of $p$ to determine the optimum value of $\lambda$ and $\tau$ for \classo and $\lambda$ for \milp (see Sec.~\ref{subsec:algorithms}).
The hyperparameter values which maximized the product of sensitivity and specificity on the validation mixture were chosen.

\begin{figure*}
    {\includegraphics[width=0.33\textwidth]{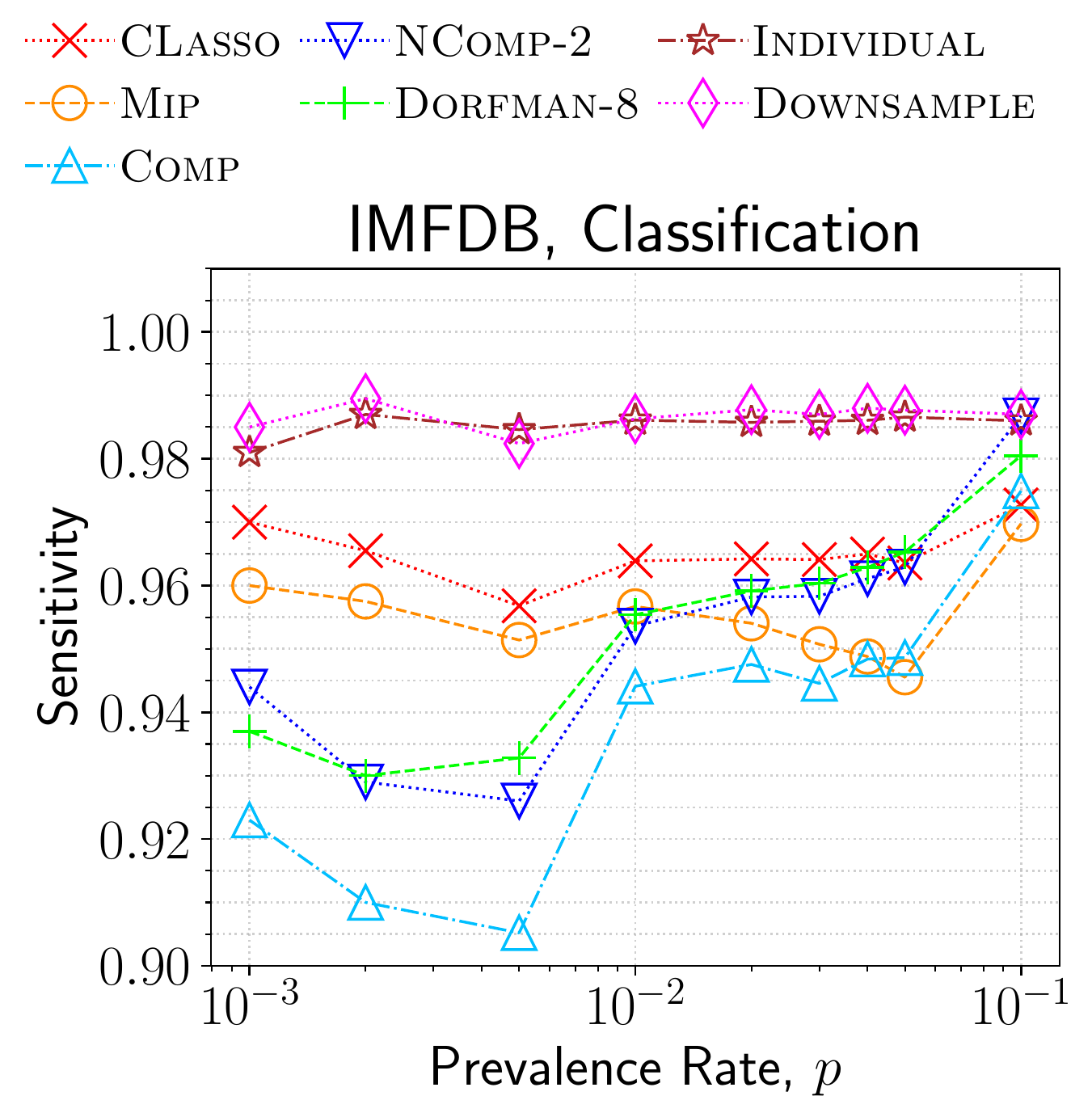}}
    {\includegraphics[width=0.33\textwidth]{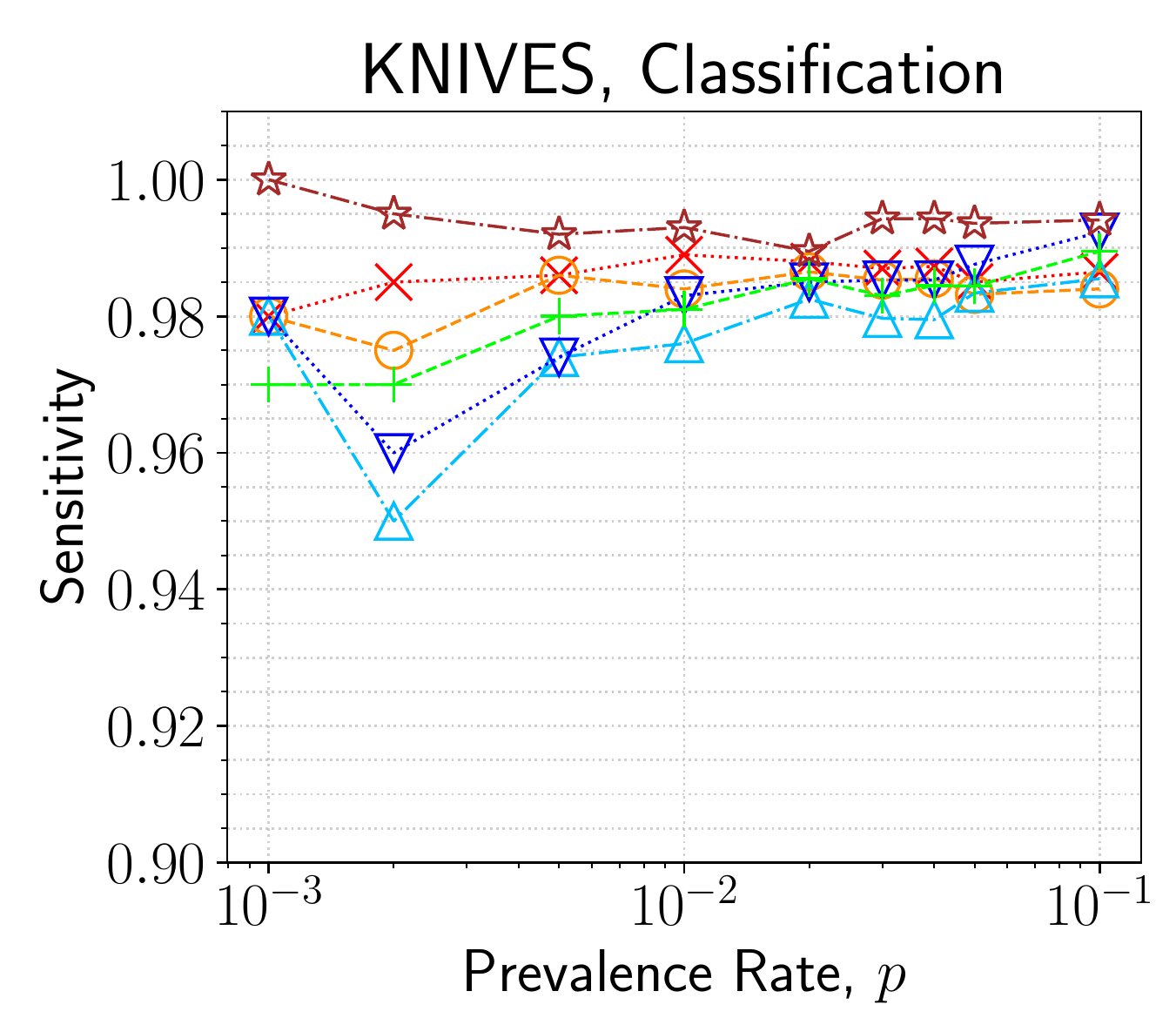}}
    {\includegraphics[width=0.33\textwidth]{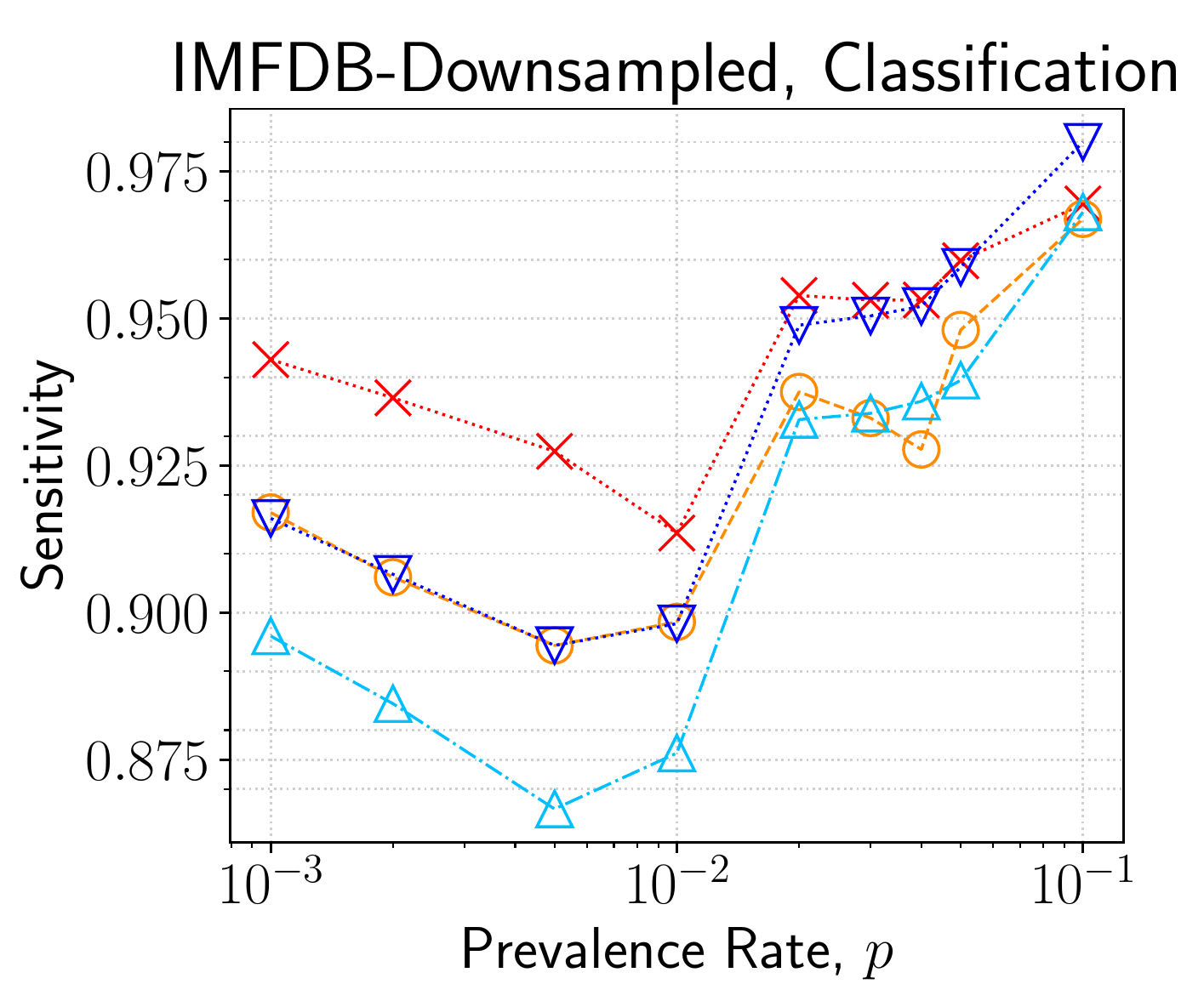}}
    \newline
    {\includegraphics[width=0.33\textwidth]{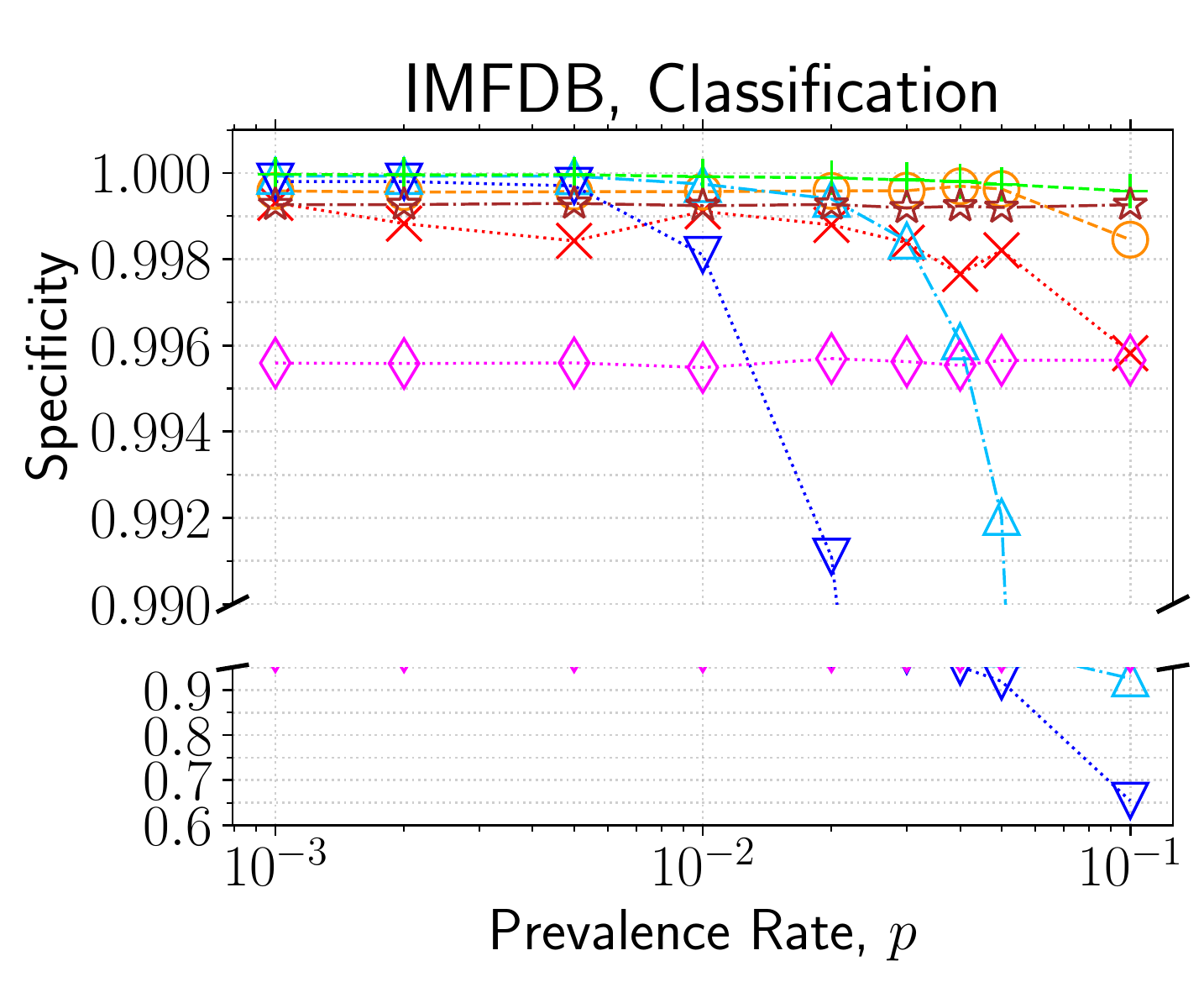}}
    {\includegraphics[width=0.33\textwidth]{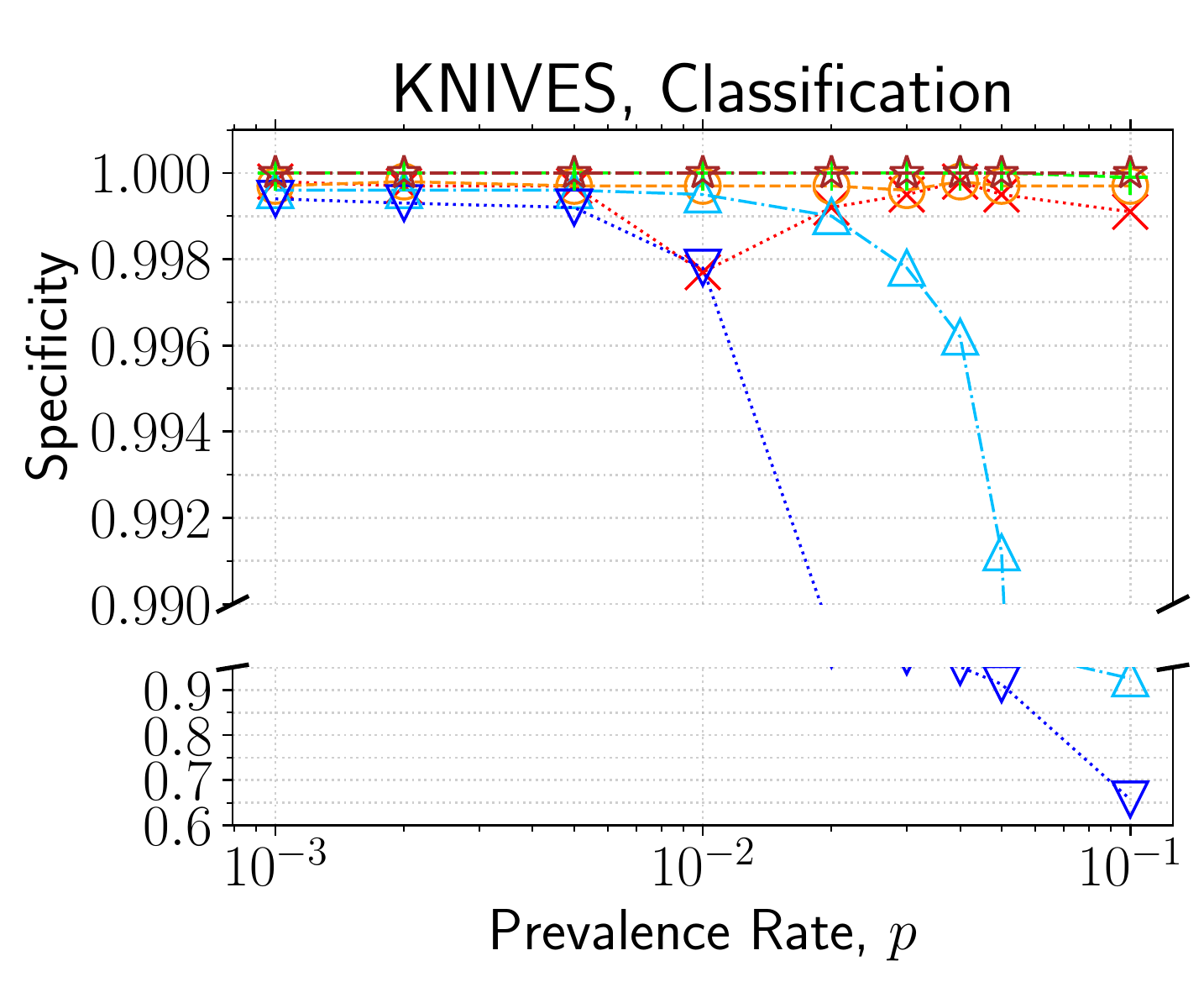}}
    {\includegraphics[width=0.33\textwidth]{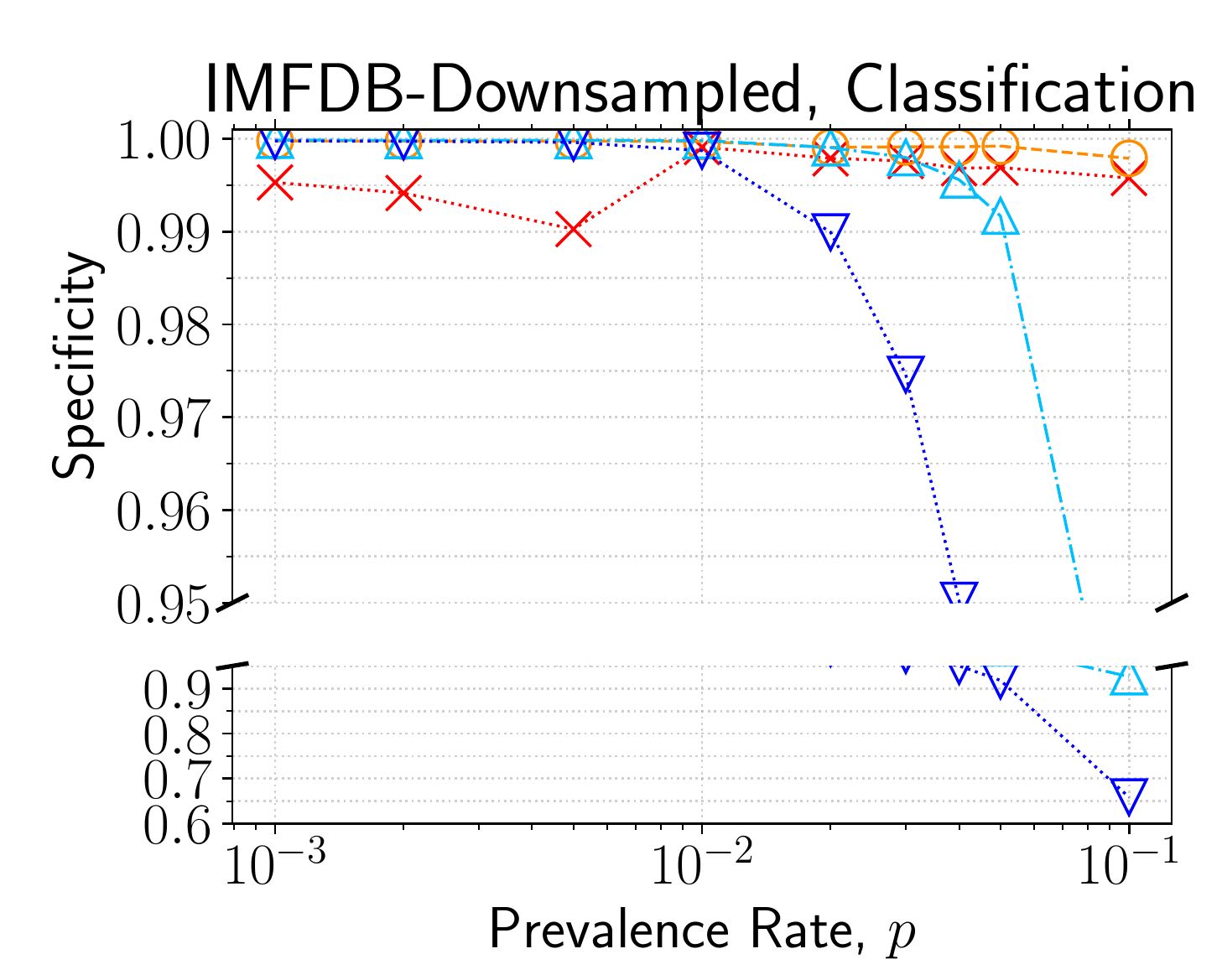}} 
    \centering
    \caption{
    Performance of Objectionable Image Moderation via Pooled Classification methods for various prevalence rates ($p$), averaged over 1M images for IMFDB and 100K images for Knives.
    Top Row: Sensitivity on IMFDB (left), Knives (middle), and IMFDB-Downsampled (right).
    Bottom Row: Specificity on IMFDB (left), Knives (middle), and IMFDB-Downsampled (right).
    Note: Y-axis for Specificity charts is broken up into two ranges in order to show all the values more clearly.}
    \label{fig:performance_p}
\end{figure*}

\paragraph{\textbf{Discussion on Results:}}
Fig.~\ref{fig:performance_p} shows the sensitivity and specificity of each of these algorithms for different values of $p$, for the two classifications tasks (\textit{i}) and \textit({ii}) defined earlier.
We see that for IMFDB and Knives,
the CS/GT methods remain competitive with \textsc{Individual}.
In general, \classo and \milp have \emph{much higher sensitivity} than \comp and \textsc{Dorfman-8}.
This is because if a pool gets falsely classified as negative (i.e. containing no OIs), then all OIs in that pool get classified as negative by \comp and \dorf.
However, \classo and \milp can recover from such errors as they inherently consider the results of other pools that the OI takes part in.
Such false negative pools are more common at lower $p$ values, where most positive pools have only a single OI, which may remain undetected by the pooled neural network (see also the confusion matrix of the pooled neural network in Fig~\ref{fig:confusion_matrix}).
\ncomp-2 improves upon the sensitivity of \comp and \dorf-8 by allowing for one false negative pool for each OI.
However, it incurs a corresponding drop in specificity because non-OIs which take part in three positive pools incorrectly get classified as OI.
When $p$ is high, it is likely that such non-OIs are common, and hence the steep drop in specificity makes \ncomp-2 unviable.
At high prevalence rates of OIs, it is common for a non-OI to test positive in all of its pools.
Since \comp uses only such binary information, it incorrectly declares all such images as OIs, and we observe a sharp decline in specificity.
However, \classo and \milp use the quantitative information of the number of OIs in the pools, and can eliminate such false positives.
For example, if it is known that a positive pool has only two OIs, then in general only two images from that pool will be declared as being an OI, whereas \comp may declare anywhere from $1$ through $8$ positives for such a pool.
Among the CS algorithms, \classo has better sensitivity, while \milp has better specificity.

\paragraph{\textbf{Comparison with Downsampling:}}
In Fig.~\ref{fig:performance_p}, we also compare our method to individual testing of images downsampled by a factor of $4$ (size $112\times 112$) (referred to in Fig.~\ref{fig:performance_p} as \textsc{Downsample}) for the firearms task. We find that downsampling lowers the specificity of prediction of individual testing, and both \milp and \classo have better specificity for both IMFDB and Knives.
\begin{figure}
    \centering
    \subfloat{
    \includegraphics[width=0.5\columnwidth]{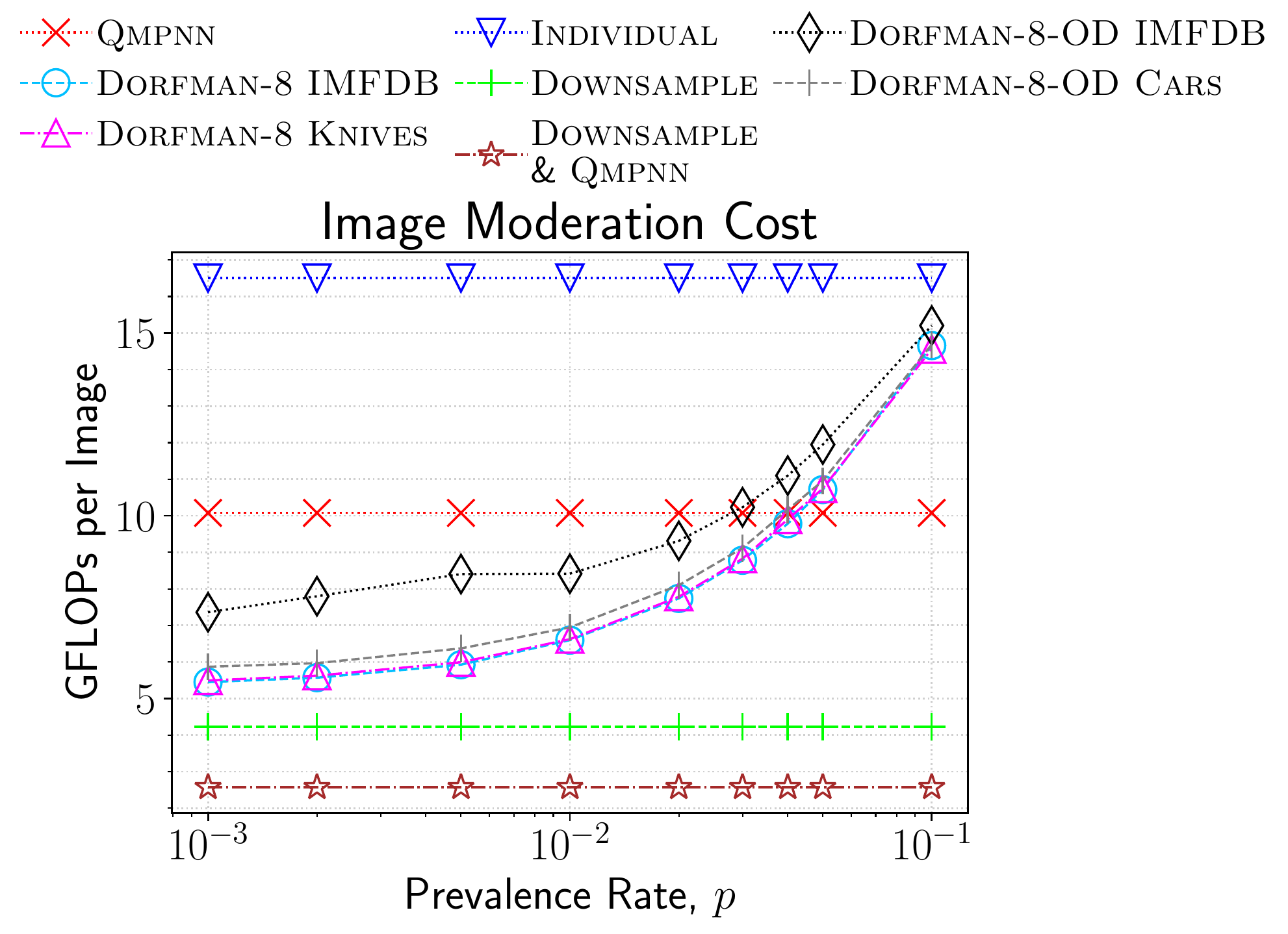}}
    \caption{Image Moderation cost per image in giga floating point operations (GFLOPs).
    Cost for running non-adaptive or individual methods is the same for classification and outlier detection.
    Cost for \dorf-8 averaged over 1M images for IMFDB and 100K images for Knives.
    Cost for \dorfod8 (see Sec.~\ref{subsec:multi_class}, paragraph titled `Prediction') averaged over 100K images for IMFDB and Cars.}
    \label{fig:flops}
\end{figure}

\paragraph{\textbf{Computational Cost:}}
Fig.~\ref{fig:flops} compares the average number of floating point operations used per image processed by the \textsc{Qmpnn} against those used for \dorf-8 and \individual for different prevalence rates on IMFDB.
We used the Python package \textit{ptflops} \cite{ptflops} for estimation of FLOPs used by the neural networks. We see that the \textsc{Qmpnn} with $50\times100$ pooling matrix uses only around $61\%$ of the computation of \individual.
While \dorf-8 uses less computation for small $p$, the \textsc{Qmpnn} is significantly more efficient for $p > 0.05$.

\paragraph{\textbf{Wall-clock Times:}}
The end-to-end wall-clock times for \Qmpnn with \milp decoding on IMFDB are presented for a range of prevalence rates in Table \ref{tab:wall_clock_times}.
We see that this is significantly faster than individual testing of the images for all $p$, and for $p \geq 0.04$ than \dorf-8.
\begin{table}
\centering
\begin{tabular}{|c|c|c|c|}
\hline
$p$ & {\textbf{Individual}} & \textbf{Dorfman-8 \cite{liang2021neural}} & \textbf{\textsc{Qmpnn} + \textsc{Mip}} \\ \hline
0.001              & \multirow{6}{*}{5899.0} & 2080.6           & 3003.5                     \\ \cline{1-1} \cline{3-4} 
0.01               &                         & 2352.7           & 2992.4                     \\ \cline{1-1} \cline{3-4} 
0.02               &                         & 2668.6           & 3007.9                     \\ \cline{1-1} \cline{3-4} 
0.03               &                         & 2924.7           & 2991.0                     \\ \cline{1-1} \cline{3-4} 
0.04               &                         & 3152.2           & 2988.1                     \\ \cline{1-1} \cline{3-4} 
0.05               &                         & 3403.6           & 2983.3                     \\ \cline{1-1} \cline{3-4} 
0.1                &                         & 4315.0           & 2984.4                     \\ \hline
\end{tabular}
\caption{End-to-end wall-clock times (secs), for processing 1M images of IMFDB+ImageNet with one NVIDIA GeForce RTX 2080 Ti GPU (11GB RAM) + one AMD
12-Core CPU, 64GB RAM. Batch-size chosen to fill  GPU RAM. 
}
\label{tab:wall_clock_times}
\end{table}

\paragraph{\textbf{Combining Pooling with Downsampling:}}
We verify whether downsampling may be combined with GT and CS methods for further reduction in computation cost without negatively affecting the performance.
Fig.~\ref{fig:performance_p} (sub-figures labelled IMFDB-Downsampled, rightmost column, top and bottom rows) shows the performance of non-adaptive GT and CS methods on IMFDB images downsampled by a factor of $4$ to $112\times 112$. We note that while all the algorithms observe a drop in sensitivity, it may be within an acceptable range depending on the particular use-case, especially for \classo. Specificity also drops, but less severely. \classo and \milp present the best balance of specificity and sensitivity across the entire range of $p$.

\paragraph{\textbf{Multiple Objectionable Elements in the \emph{Same Image:}}}
We created $100$ random pools containing one OI (firearm image) each, but where \textit{each OI depicted two firearms}.
Out of $100$, the \textsc{Qpnn} correctly reported $97$ times that the pool contained only one OI, and incorrectly reported two OIs $3$ times.
We conjecture that the \textsc{Qpnn} detects unnatural edges between firearms in the SFM, due to which it does not get confused by a single image containing multiple firearms.

\subsection{Off-topic Image Moderation using Pooled Outlier Detection}
\label{subsec:multi_class}
Here, the on-topic images are from a single underlying class and off-topic images may be from classes not seen during training. 
The goal is to detect such off-topic images, which do not belong to the underlying class.
We do this by treating this as an outlier detection task, where the on-topic images are considered `normal' and off-topic images are considered to be outliers.
\paragraph{\textbf{Tasks:}}
We consider two off-topic image moderation tasks, wherein the on-topic images are (\textit{i}) firearm images from IMFDB used in Sec.~\ref{subsec:single_class}, and (\textit{ii}) car images from the Stanford Cars Dataset \cite{StanfordCars}.  
During training, we use non-firearm or non-car images from ImageNet-1K as off-topic images from known classes for the two tasks, respectively.
During training, we use images from some 182 non-firearm or 179 non-car random classes which are in ImageNet-21K \cite{deng2009imagenet,ridnik2021imagenet} but not in ImageNet-1K.
The off-topic image classes chosen for testing are not present in the training datasets.
\paragraph{\textbf{Individual Dataset Splits:}}
The firearm images in training, validation and test splits of IMFDB as described in Sec.~\ref{subsec:single_class} are taken to be on-topic images for the respective splits for the off-topic image moderation task. From the 976 non-firearm classes of ImageNet-1K in Sec. ~\ref{subsec:single_class}, we randomly sample \num{10000} and \num{1000} images and add them to the training and validation split, respectively, as off-topic images.
For testing, we choose \num{1000} images from \num{182} random non-firearm classes of ImageNet-21K as off-topic images.

For the Cars dataset, we take 8041 cars images for testing, 2144 cars images for validation, and 6000 cars images for training from the Stanford cars dataset as on-topic images.
The \num{1000} off-topic images for testing come from 179 non-car classes of ImageNet-21K, whereas for training and validation, we sample 10000 and 1000 images from 955 non-cars classes of Imagenet-1K. 
We created $9$ test sample sets (one per prevalence rate) of the test data split, each containing 100K images, using the same procedure and the same prevalence rates $p$ as in Sec.~\ref{subsec:single_class}.
At each epoch of \textsc{Inn} training, classes in the training split were balanced by using all \num{10000} off-topic images from known classes and randomly sampling (with replacement) an equal number of on-topic images.
The validation split classes were not balanced since the imbalance in class proportions was not too much.

\paragraph{\textbf{Pooled Dataset Splits:}}
The pooled dataset training split was created using the same procedure as in Sec.~\ref{subsec:single_class}, with the same pool size (i.e. 8), number of pools, and distribution of pools with different number of off-topic images in them.
The pooled dataset validation split had \num{625} pools and was also created using the same procedure.

\paragraph{\textbf{Neural Network, Data Transformation, Training:}}
We use the same neural network (ResNeXt-101 \cite{pytorchresnext,xie2017resnext}) and train an \textsc{Inn} and a \textsc{Qpnn} using the same procedure as 
for the classification case (Sec.~\ref{subsec:single_class} and Sec.~\ref{subsec:pooled_nn}) on the IMFDB and Cars datasets.
The data transformation used was the same as for the classification case, except that random rotation transformations were also applied to the images during testing to create more artificial edges in the SFMs and aid in detection of outliers in a pool.

\begin{figure}
    \centering
    \includegraphics[width=0.5\textwidth]{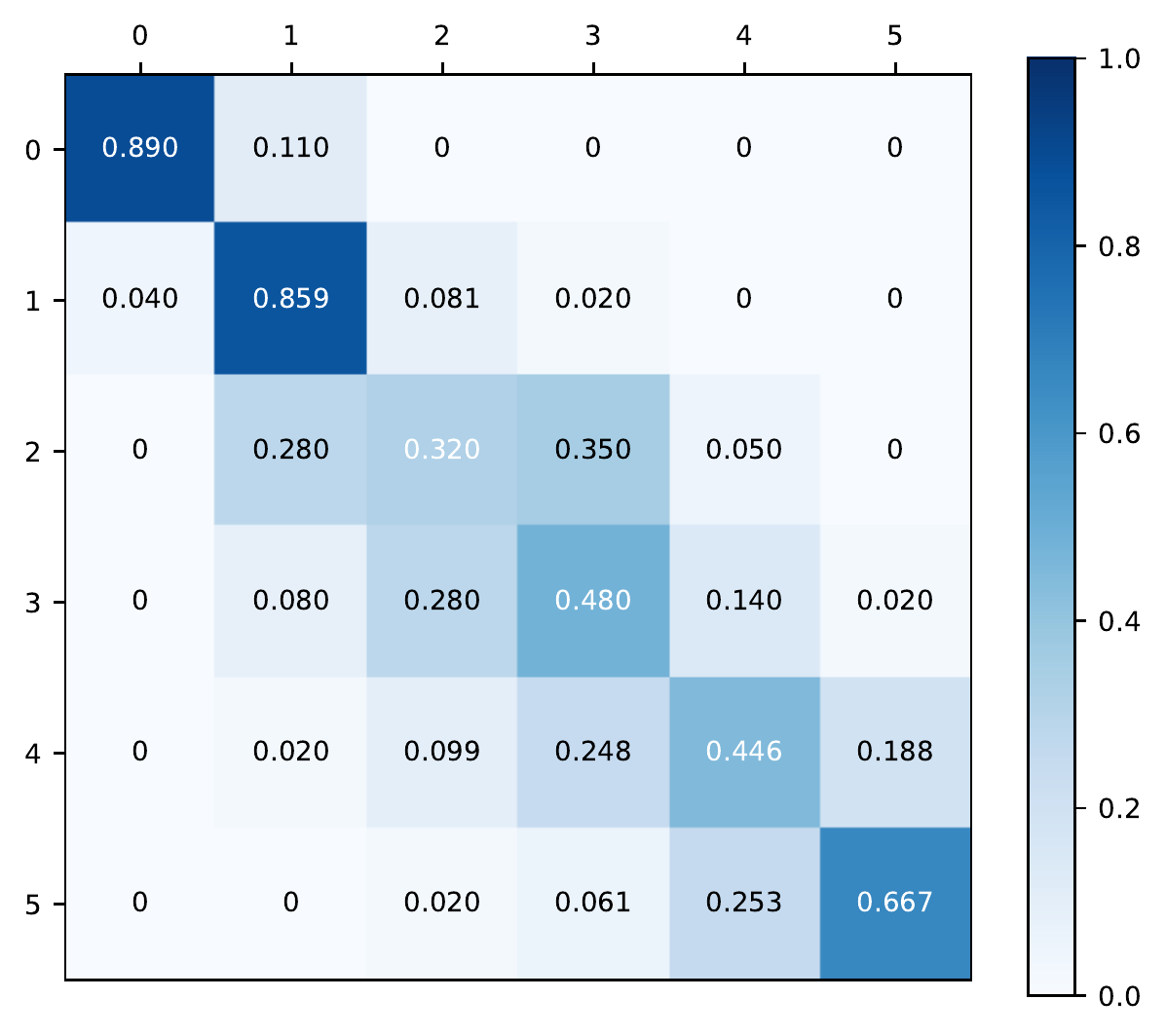}
    \caption{Histogram method Confusion Matrix on IMFDB Pooled Validation Data: true \#OIs (rows) versus predicted \#OIs (columns)}
    \label{fig:confusion_matrix_val_ood}
\end{figure}

\paragraph{\textbf{GMM and Histogram Creation:}}
Recall the steps for GMM fitting and creation of histograms of anomaly scores in Sec.~\ref{subsec:outlierdetection}.
We created $M_{\text{IMFDB}} = 40850$ and $M_{\text{Cars}} = 18750$ pools respectively from the training splits of IMFDB and Cars datasets and obtained their feature vectors by passing them to the trained ResNeXt-101 \Qpnn.
For each dataset, we fit a GMM on these feature vectors.
The optimal number of clusters in the GMM for the pooled case was 30 and 6 respectively for the IMFDB and Cars dataset, and 5 and 1 respectively for the individual case.
For computing the anomaly scores histograms for each dataset, a total of $N = 100K$ pools were created from the validation split, each of size $r=8$ and maximum number of outliers $t=5$ (roughly 16666 pools for each pool label).
We capped the number of outliers at $5$ because for small prevalence rate $p$ ranging between 0.001 and 0.1, having more than 5 outlier images in a pool of 8 is a very low probability event.
For each dataset, we passed these pools to the trained \Qpnn to obtain pool feature vectors. We used the earlier GMM to obtain their anomaly scores, and computed the anomaly score histogram, with $Q = 500$ bins.
Recall from Sec.~\ref{subsec:outlierdetection} and Algorithm~\ref{alg:ood_dorfman} that the same histogram method can be used for testing individual images as well. For creating the histogram for testing individual images, we used 5000 off-topic and 5000 on-topic images, and used  $Q=500$ bins.

\paragraph{\textbf{Confusion Matrix:}}
Fig.~\ref{fig:confusion_matrix_val_ood} presents the $(t+1) \times (t+1)$ confusion matrix of the ground truth number of OIs in a pool versus the number predicted by the histogram method.
The method predicts upto $t=5$ outlier images in a pool.
We see that if $g$ is the ground truth number of OIs, the predictions of our proposed method lie in the interval $[\textrm{max}(0,g-1),\textrm{min}(g+1,t)]$ with high probability.

\paragraph{\textbf{Pooling Matrix:}}
The same $50 \times 100$ balanced pooling matrix $\boldsymbol{\Phi}$ as in the pooled classification case (Sec.~\ref{subsec:single_class}) is used for pooled outlier detection as well.

\paragraph{\textbf{Prediction:}}
For each dataset, we created $9$ test sample sets (one per prevalence rate) each of size 100K with different prevalence rates of off-topic images, similar to Sec.~\ref{subsec:single_class}.
For pooled outlier detection using CS or non-adaptive GT methods in Sec.~\ref{subsec:algorithms}, Algorithm~\ref{alg:ood_inference} was used, with the $50\times100$ pooling matrix $\boldsymbol{\Phi}$, processing $n = 100$ images at a time.
For Dorfman pooled outlier detection, Algorithm~\ref{alg:ood_dorfman} was used, processing $r = 8$ images at a time.
We refer to these outlier detection methods as \classood, \compod, \ncompod{2}, and \dorfod{8} in the discussion that follows.
Individual testing using our outlier detection method is referred to as \individualod, and individual testing of downsampled images as \downsampleod.

\begin{figure*}
    {\includegraphics[width=0.37\textwidth]{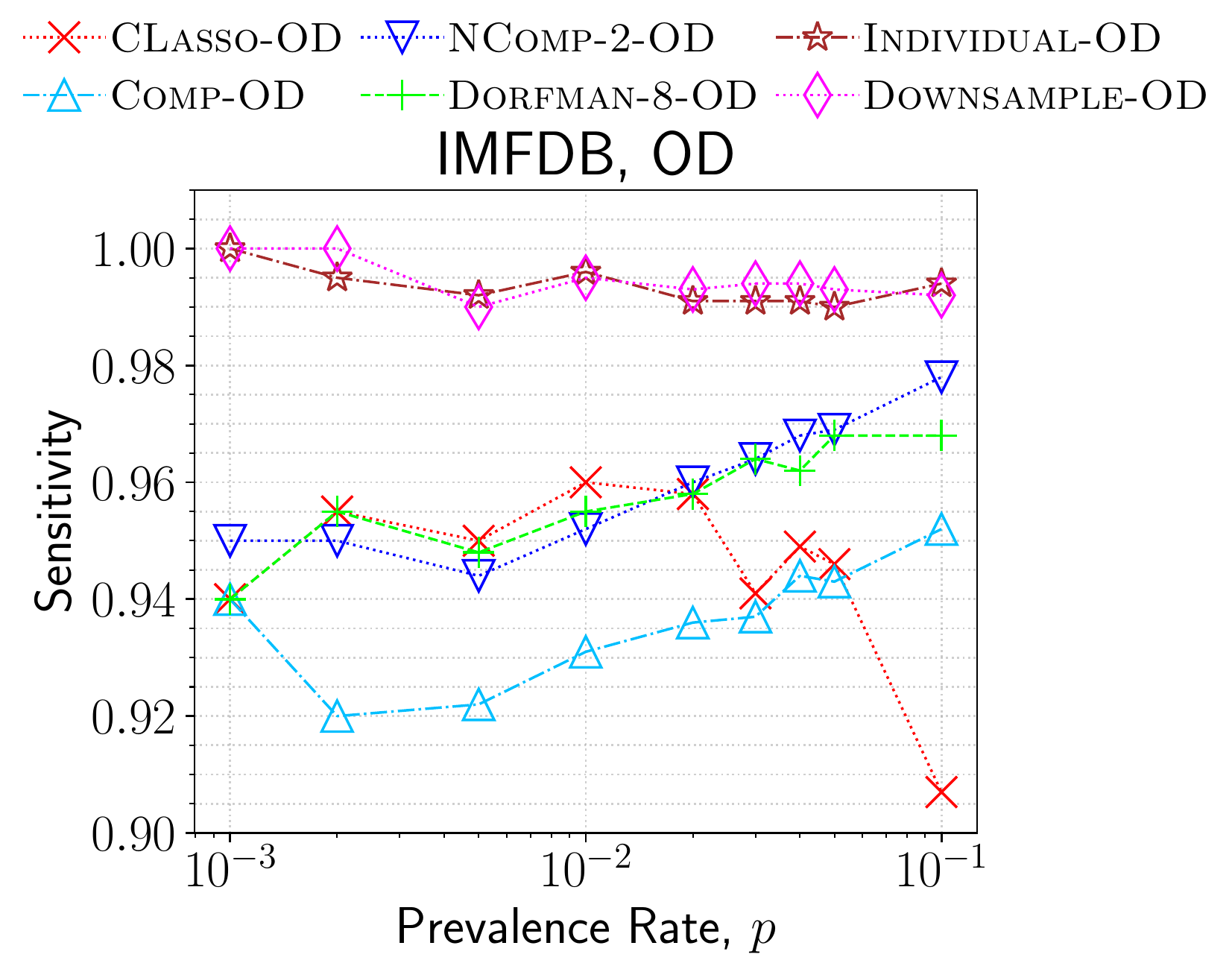}}
    {\includegraphics[width=0.30\textwidth]{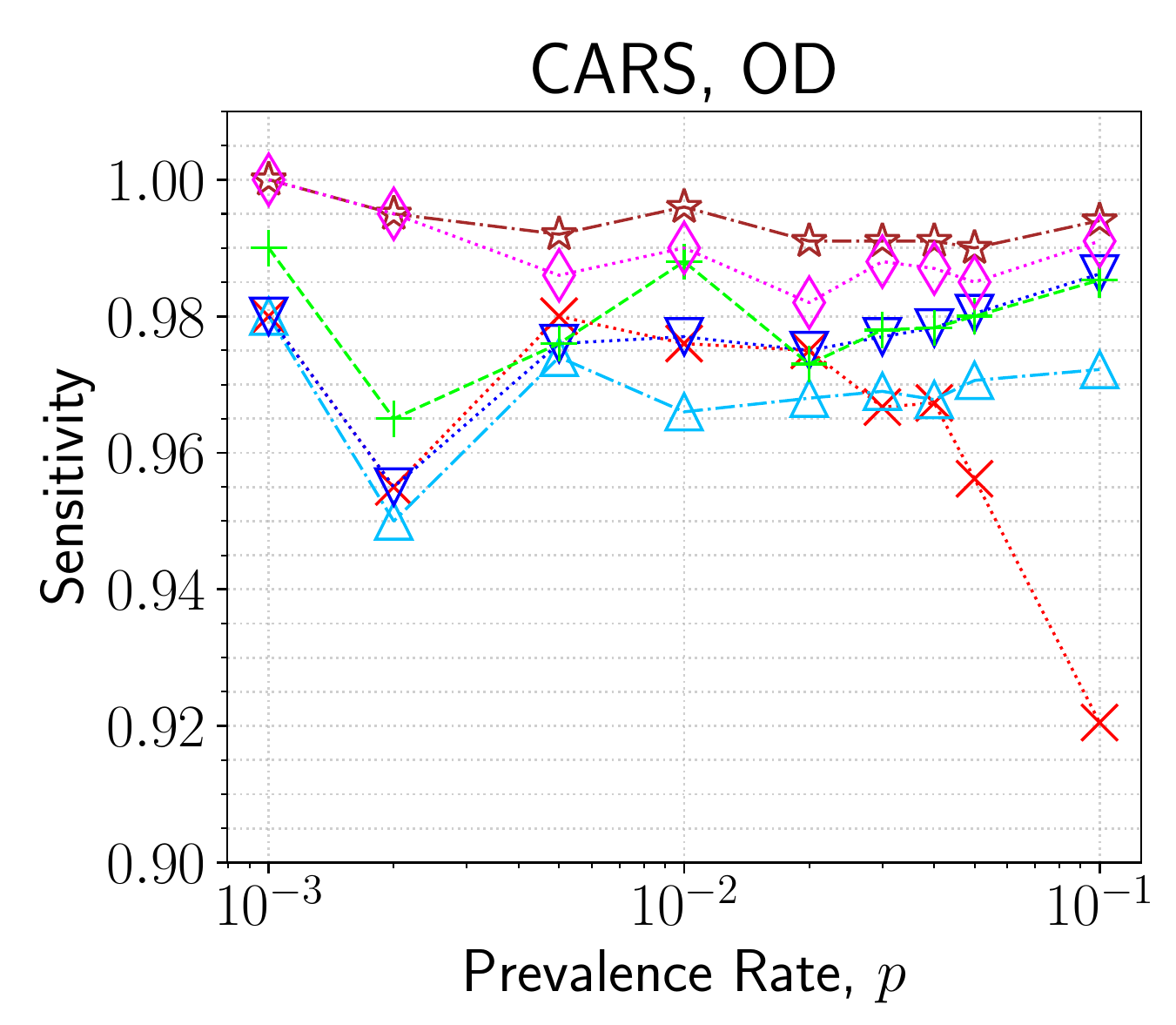}}
    {\includegraphics[width=0.30\textwidth]{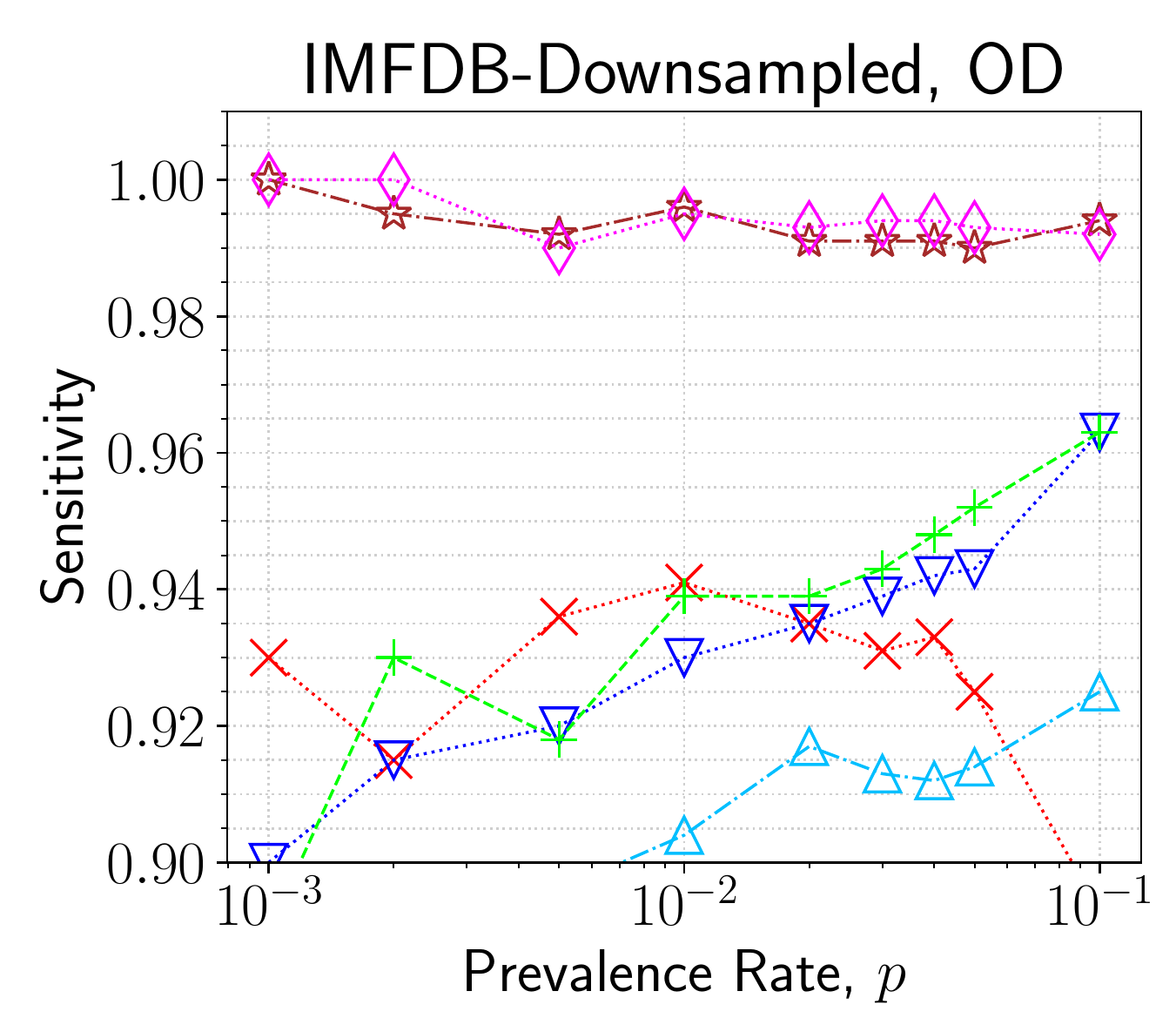}}
    \newline
    {\includegraphics[width=0.33\textwidth]{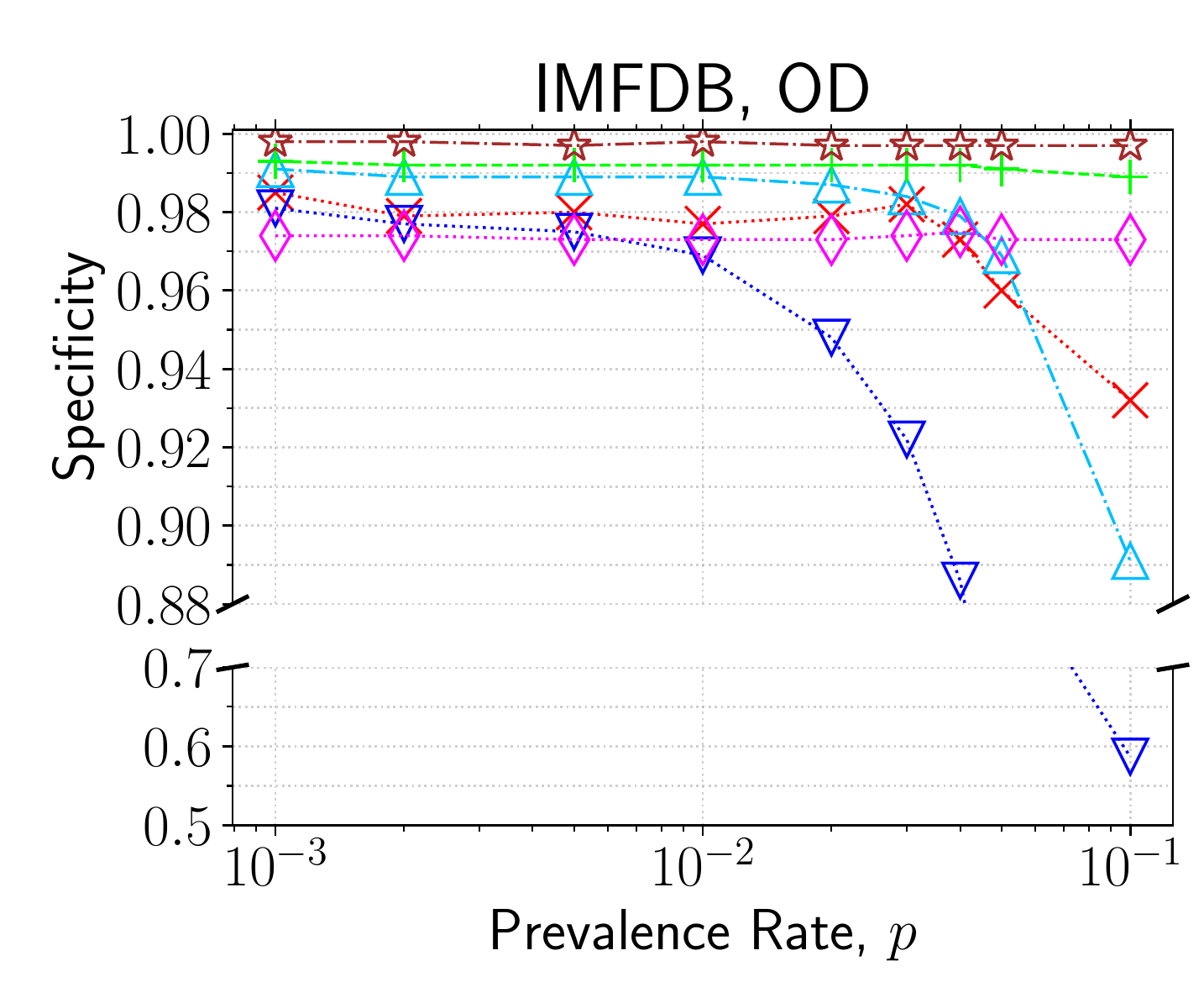}}
    {\includegraphics[width=0.33\textwidth]{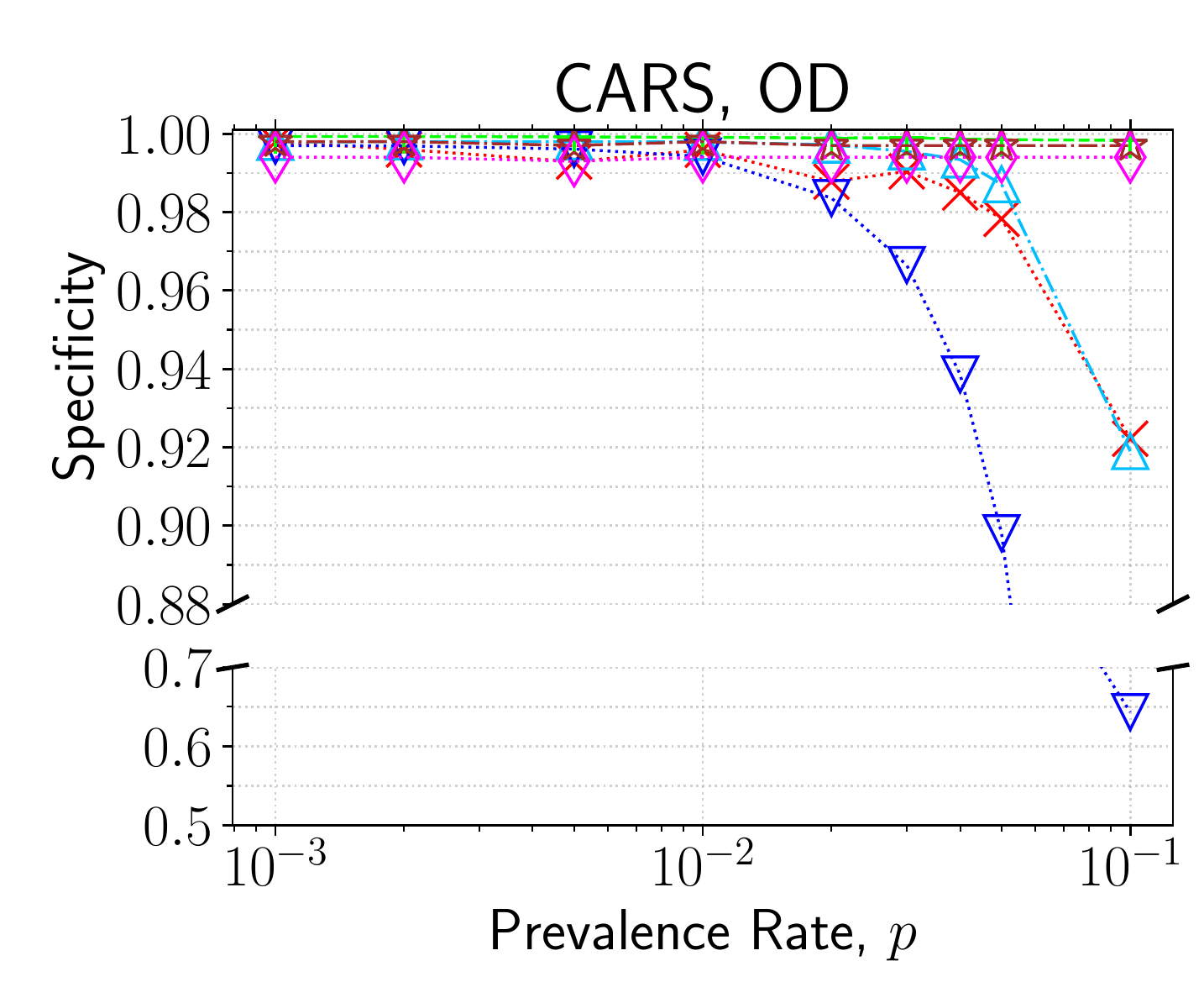}}
    {\includegraphics[width=0.33\textwidth]{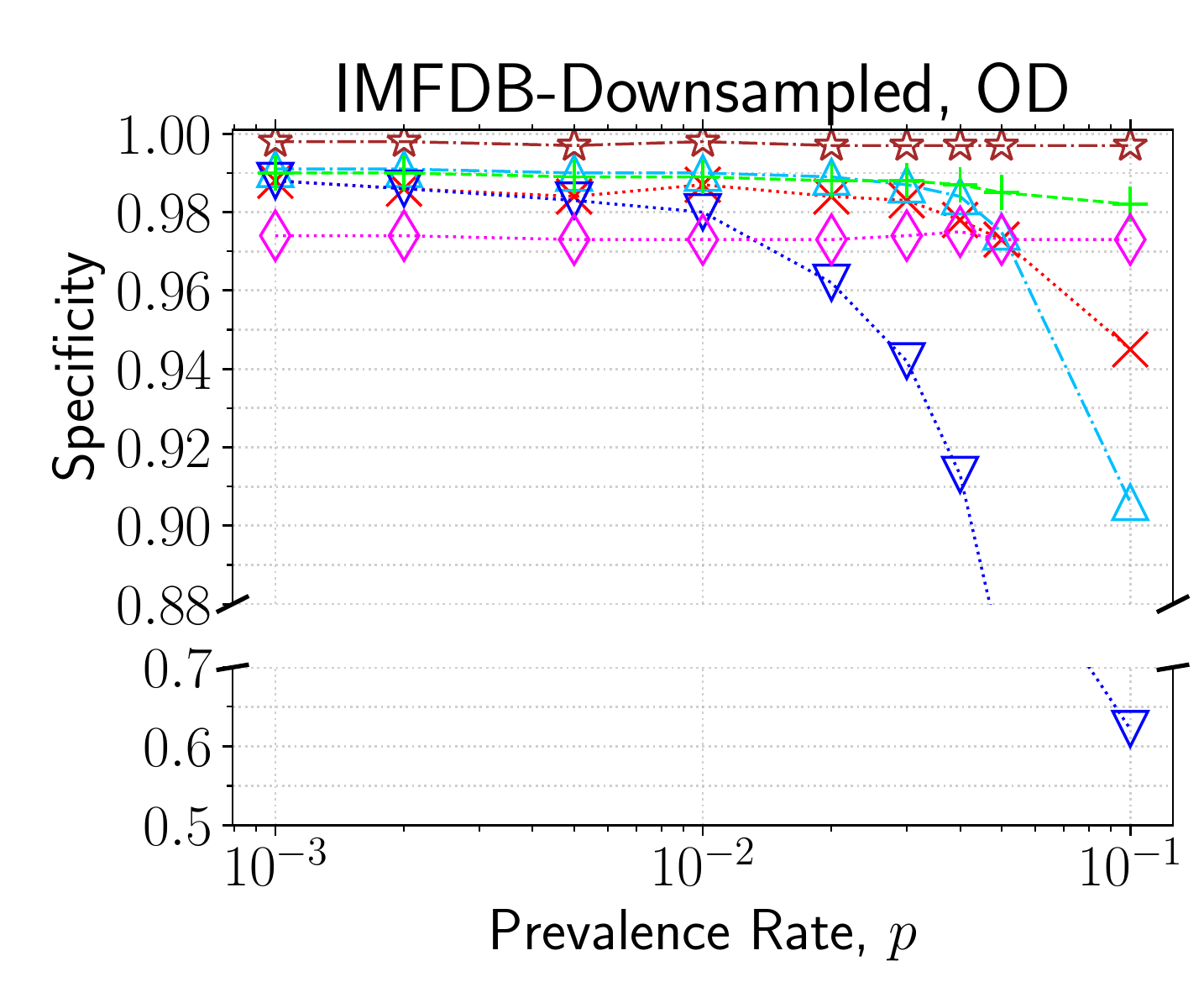}} 
    \centering
    \caption{Performance of Off-topic Image Moderation via Pooled Outlier Detection methods
    for various prevalence rates ($p$), averaged over 1M images for IMFDB and 100K images for Cars.
    Top Row: Sensitivity on IMFDB (left), Cars (middle), and IMFDB-Downsampled (right).
    Bottom Row: Specificity on IMFDB (left), Cars (middle), and IMFDB-Downsampled (right). 
    Note: Y-axis for Specificity charts is broken up into two ranges in order to show all the values more clearly.}
    \label{fig:performance_p_ood}
\end{figure*}
\paragraph{\textbf{Hyperparameter Selection:}}
The hyperparameters for \classood were chosen in exactly the same manner as for the classification case (Sec.~\ref{subsec:single_class}).

\paragraph{\textbf{Discussion on Results:}}
Fig.~\ref{fig:performance_p_ood} shows the sensitivity and specificity of each of these algorithms for different values of $p$, for the two off-topic image detection tasks defined earlier.
We see that for both IMFDB and Cars dataset,
the CS/GT methods for oulier detection remain competitive with \textsc{Individual}.
We observe that \classood has higher sensitivity than \compod for the entire range of prevalence rates considered and higher than \dorfod{8} for $p\leq0.01$.
The reason for this has been discussed in Sec.~\ref{subsec:single_class}: when a pool gets \emph{falsely} classified as negative (i.e., containing no outlier image), then all outlier images in that pool get classified as negative by \compod and \dorfod{8}.
In the higher $p$ regions, \dorfod{8} starts to perform better than \classood. This is because, as we can see in Fig.~\ref{fig:confusion_matrix_val_ood}, the prediction model gets confused for pools having $\geq$ 2 outlier images.
At high prevalence rates of off-topic images, it is common for an on-topic image to test positive in all of its pools.
As discussed in Sec.~\ref{subsec:single_class}, \compod uses only such binary information and declares incorrectly all such images as off-topic, so we observe a sharp decline in specificity.
However, \classood uses the quantitative information of the number of off-topic images in the pools, and can eliminate such false positives.

\paragraph{\textbf{Comparison with Downsampling:}}
In Fig.~\ref{fig:performance_p_ood}, we also compare our method to individual testing of images downsampled by a factor of $4$ (size $112\times 112$) (referred to in Fig.~\ref{fig:performance_p_ood} as \downsampleod) for the firearms task.
We find that downsampling lowers the specificity of prediction and hence both \classood and \dorfod8 show better specificity than individual testing of downsampled images.

\paragraph{\textbf{Computational Cost:}}
The computational cost in terms of number of floating point operations (FLOPs) are presented in Fig.~\ref{fig:flops}. 
Since the same neural network as in the classification case has been used, the amount of computation for each image for pooled outlier detection using CS/non-adaptive GT methods is the same. Some additional computation is needed in the GMM related steps, but it is very small (by a factor of 100) compared to computations involving the neural network.
For \dorfod8, the amount of computation needed is slightly higher than for \dorf-8.
This is because in case of pooled outlier detection, $11\%$ of negative pools falsely test as positive, as compared to only $2\%$ for pooled classification (see the top-left entries in the confusion matrices presented in Fig.~\ref{fig:confusion_matrix_val_ood} and Fig.~\ref{fig:confusion_matrix}).
However, we see that the computation costs for both adaptive and non-adaptive pooled outlier detection methods are well below those of individual testing.

\paragraph{\textbf{Wall-clock Times:}}
The end-to-end wall-clock times for pooled outlier detection with \classo decoding and using \dorfod{8} on the Cars Dataset for 100K images are presented for a range of prevalence rates in Table~\ref{tab:wall_clock_times_2}.
We see that both \classood and \dorfod{8} are significantly faster than individual testing of the images for all $p$.
\classood is also faster than \dorfod8 for $p \geq 0.02$.

\paragraph{\textbf{Combining Pooling with Downsampling:}}
Similar to Sec.~\ref{subsec:single_class}, we performed experiments to see whether downsampling can be combined with GT and CS methods for further reduction in computation cost without negatively affecting the performance.
We show the performance of our pooled outlier detection methods on IMFDB images downsampled by a factor of $4$ to $112\times 112$ in Fig.~\ref{fig:performance_p_ood} (sub-figures labelled IMFDB-Downsampled, rightmost column, top and bottom row).
We again note that while all the algorithms observe some drop in sensitivity, it is within an acceptable range. The specificity also drops, but to an even smaller extent.
\begin{table}
\centering
\begin{tabular}{|c|c|c|c|}
\hline
$p$ & {\textbf{\textsc{Individual-OD}}} & \textbf{\dorfod8} & \textbf{\textsc{Classo-OD}} \\ \hline
0.001              & 430.87      &      253.6    &      302.2         \\ \hline 
0.01               &       437.5                  &           281.6    &     305.6              \\ \hline 
0.02               &            436.9                &            309.5      &     303.3        \\ \hline 
0.03               &            437.3               &            332.9   &      304.1          \\ \hline 
0.04               &             437.5                  &           356.9     &      303.4      \\ \hline 
0.05               &             436.7              &          376.0      &       303.2         \\ \hline 
0.1                &            436.5               &            466.3      &      303.4        \\ \hline
\end{tabular}
\caption{End-to-end wall-clock times (secs), for processing 100k images of Cars+ImageNet with one NVIDIA GeForce RTX 2080 Ti GPU (11GB RAM) + one AMD 
12-Core CPU, 64GB RAM. Batch-size chosen to fill  GPU RAM. 
}
\label{tab:wall_clock_times_2}
\end{table}

\section{Conclusion and Future Work}
We presented a novel CS based approach for efficient automated image moderation, achieving significant reduction in computational cost for image moderation using deep neural networks.
We also presented a method for pooled deep outlier detection for off-topic image moderation, bringing for the first time CS and GT methods to outlier detection.
Because we output the number of OIs for each pool, recovering the original OIs is also a quantitative group testing (QGT) \cite{Scarlett2017, Gebhard2019} problem.
Hence QGT algorithms (eg: Algorithms 1 and 2 from \cite{Gebhard2019}) may also be used instead of CS for decoding within our framework.
We leave an exploration of using QGT algorithms for image moderation as future work.
In this work, we assumed that the errors in the different elements of $\boldsymbol{y}$ are independent of each other and the underlying $\boldsymbol{x}$. Though this produced good results, using a more sophisticated noise model is an avenue for future work. Our approach reduces the number of tests and is orthogonal to approaches such as network pruning or weights quantization which reduce the complexity of each test. In future, we could combine these two approaches, just as we (already) explored the combination of pooled testing and image downsampling. 

\appendix
\section{Appendix}
\subsection{Properties of Good Sensing Matrices}
\label{sec:properties_sensingmatrices}
In a \textbf{$k$-disjunct pooling matrix}, the support of any column is not a subset of the union of supports of any $k$ other columns. In noiseless binary group testing, \textsc{Comp} can identify upto $k$ defectives exactly if the pooling matrix is $k$-disjunct \cite[Sec 2.2]{Mazumdar2012}.

\textbf{Mutual Coherence} of a matrix is the maximum value of the dot product of any of its two columns. Sensing matrices with low mutual coherence are preferred for compressed sensing as they allow for reduction of the upper bounds on recovery errors \cite[Theorem 1]{Studer2014}.

The \textbf{$1$-norm Restricted Isometry Property (RIP-1)} of order $2k$ is a sufficient condition on the sensing matrix $\bPhi$ for recovery of $k$-sparse $\x$ from noisy $\y$ via LASSO \cite{berinde2008combining}.
A pooling matrix $\bPhi$ satisfies RIP-1 of order $2k$ if it holds for some constants $2k$ and $\delta$ that $\|\x\|_1 \leq \|\bPhi \x\|_1 \leq (1+\delta)\|\x\|_1$ for all $2k$-sparse vectors $\x$.

\subsection{Choice of Pooling Matrix $\boldsymbol{\Phi}$}
\label{sec:pooling_matrix}
Randomly generated matrices from zero-mean Gaussian or Rademacher distributions have been popular in CS because they are known to obey the Restricted Isometry Property (RIP) with high probability \cite{Baraniuk2008}, which is a well-known sufficient condition to guarantee accurate recovery \cite{Candes2008b}. RIP-obeying matrices also satisfy condition $\mathscr{C}2$ mentioned in Sec. 1 of the main paper. Despite this, such matrices are not suitable for our image moderation application, as they will lead to all $n$ images contributing to every SFM -- that too in unequal or both positive and negative amounts -- and complicate the job of the \textsc{Qmpnn} that predicts $\boldsymbol{y}$.
Random Bernoulli ($\{0,1\}$) matrices have been known to allow for very good recovery \cite{kueng2017robust} due to their favorable null-space properties.
But randomly generated matrices will contain different number of ones in each row, due to which each pool would contain contributions from a different number of images.
Instead, we use binary matrices which are constrained to have an equal number of ones in each of the $n$ columns, denoted by $c$ (i.e. `column weight' $c$), and an equal number of ones in each of the $m$ rows, denoted by $r$ (i.e. `row weight' $r$).
This implies that $nc = mr$.
Such a construction ensures that the \textsc{Qmpnn} must produce outputs constrained to lie in $\{0,1,...,r\}$, and also ensures that each of the $n$ images contributes to at least one pool (actually to exactly $c$ different pools, to be precise).
These matrices can be made very sparse ($r \ll n$), which ensures that noise in the output of the \textsc{Qmpnn} can be controlled.
We put an additional constraint, that the dot product of any two rows must be at most $1$, and similarly the dot product of any two columns must be at most $1$.
It can be shown that such matrices are $(c-1)$-disjunct, obey the `RIP-1' property of order $2c$ \cite[Sec. III.F]{Ghosh2021}, and have low mutual coherence \cite{Goenka2021}, which makes them beneficial for GT as well as CS based recovery. These properties are defined in Sec.~\ref{sec:properties_sensingmatrices}.

\subsection{Recovery Guarantees for CS and binary GT}
\label{sec:recovery_guarantees}
Quantitative group testing (see Sec.~\ref{sec:background}), which is closely related to CS, has some conceptual advantages over binary group testing, as it uses more information inherently.
We summarize arguments from \cite[Sec. 2.1]{Mazumdar2012}, \cite[Sec. 1.1]{Gebhard2019} regarding this: Consider binary group testing with $m$ pools from $n$ items with $k$ defectives.
The total number of outputs is $2^m$.
The total number of ways in which up to $k$ defective items can be chosen from $n$ is $\sum_{i=0}^k C(n,i)$.
We must have $2^m \geq \sum_{i=0}^k C(n,i)$, which produces $m \geq k \log (n/k)$.
If we instead consider quantitative group testing, then the output of each test is an integer in $\{0,1,...,k\}$. The total number of outputs would thus be $(k+1)^m$. Thus $(k+1)^m \geq \sum_{i=0}^k C(n,i)$, which produces $m \geq \dfrac{k \log (n/k)}{\log (k+1)}$. Thus, the lower bound on the required number of tests is smaller in the case of quantitative group testing by a factor of $\log (k+1)$. Moreover, work in \cite{Aldridge2019} argues that in the linear regime where $k = O(n)$ (even if $k \ll n$), individual testing is the optimal scheme for binary non-adaptive group testing. This is stated in Theorem 1 of \cite{Aldridge2019} which argues that if the number of tests is less than $n$, then the error probability is bounded away from 0.
On the other hand CS recovery with binary matrices is possible with $m = O( k \log (n/k))$ measurements \cite[Theorem 9]{berinde2008combining}.

\subsection{Upper bound on disjunctness of column-regular matrices}
\label{sec:disjunctness_upper_bound}
We explain in this section that matrices with column weight $k$ (i.e., with $k$ ones per column, equivalent with each image participating in $k$ pools) which have strictly fewer rows than columns cannot be $k$-disjunct.
The work in \cite[Prop. 2.1]{erdos1985families} gives an upper bound on the cardinality of a uniform $r$-cover-free family of sets.
They consider the power set of a set $X$ of $n$ elements.
A $k$-uniform family of sets is any set of subsets of $X$ containing exactly $k$ elements.
Let $\mathcal{F} = \{S_1,\dots,S_K\}$ be a $k$-uniform family of sets of cardinality $K$.
If $\mathcal{F}$ is $r$-cover-free, it means that no element of $\mathcal{F}$ is a subset of the union of any $r$ other sets in $\mathcal{F}$.
The work in \cite[Prop. 2.1]{erdos1985families} gives the following bound on the cardinality of $F$:

\begin{equation}
\label{eq:}
    K \leq \frac{{n \choose t}}{{{k-1}\choose{t-1}}},
\end{equation}
where $t = \lceil {\frac{k}{r}} \rceil$.

We can construct a $n\times K$ pooling matrix $\boldsymbol\Phi$ from $\mathcal{F}$, where each column is a $0/1$-binary vector with column weight $k$, and $\boldsymbol\Phi_{ij} = 1$ indicating that the element $i$ is in the set $S_j$.
Recall from Appendix~\ref{sec:properties_sensingmatrices} that in an $r$-disjunct matrix, the support of any column is not a subset of the union of supports of any $r$ other columns.
From the definition of disjunctness and $r$-cover-free families, if the matrix $\boldsymbol{\Phi}$ is to be $r$-disjunct, then the family of sets $\mathcal{F}$ must be $r$-cover free.
Moreover, if $r=k$, then $t=1$, and $K \leq \frac{n}{k-1}$.
Hence, matrices with column weight $k$ and fewer rows than columns i.e. $n < K$ cannot be $k$-disjunct.

For the case from Sec.~\ref{subsec:differences} where column weight can be \emph{upto} $k=2$, we note that for any column with column weight $1$, the corresponding row in which it has a $1$ entry also has a row weight equal to $1$, otherwise the matrix cannot be $2$-disjunct.
Thus, we can remove such rows and columns from the matrix, and the reduced matrix will also be $2$-disjunct and will have column weight equal to $2$ for all columns.
From the above result, the number of columns in this reduced matrix must be less than or equal to the number of rows.
Since we removed an equal number of rows and columns from the original matrix, the original matrix also has number of columns less than or equal to the number of rows, and hence cannot be used for achieving a reduction in number of tests using group testing.

\subsection{Noise-tolerance of balanced binary matrices with row and column dot product at most 1}
\label{sec:binary_matrices_noise_tolerance}
The work in \cite[Sec. III.F.4]{Ghosh2021} shows that such matrices with column weight $c$ are adjacency matrices of $(k, \epsilon)$-unbalanced expander graphs, and consequently \cite[Theorem 1]{berinde2008combining} have RIP-1 for any $k < 2c+1$, with $\epsilon = \frac{k-1}{2c}$, and $0 \leq \epsilon < 1$.
Robust recovery guarantees using the RIP-1 for such adjacency matrices exist -- for example, \cite[Theorem 6.1]{khajehnejad2011sparse} states that the $\ell_1$-norm of the recovery error of \cite[Algorithm 1]{khajehnejad2011sparse} is upper-bounded by $\frac{7-4\epsilon}{1-2\epsilon}\eta$, where $\eta$ is the $\ell_1$-norm of the noise in measurements.
For a fixed magnitude of noise in the measurements, the amount of error in the recovered vector depends on the value of $\epsilon$, with smaller values being better.
For a fixed $k$, $\epsilon$ can be made smaller by having a larger value of $c$, the column weight of the matrix.
Hence matrices with higher values of the column weight i.e. ones in which each item takes part in a larger number of pools are more robust to noise for CS recovery.

\bibliographystyle{abbrvnat}

\bibliography{egbib}

\end{document}